\documentclass{article}

\PassOptionsToPackage{numbers,sort&compress}{natbib}
\usepackage[final,eandd]{neurips_2026}
\usepackage{graphicx}

\usepackage[utf8]{inputenc} 
\usepackage[T1]{fontenc}    
\usepackage{hyperref}       
\usepackage{url}            
\usepackage{booktabs}       
\usepackage{amsfonts}       
\usepackage{nicefrac}       
\usepackage{microtype}      
\usepackage[table]{xcolor}         
\usepackage{siunitx}
\usepackage{multirow}  
\usepackage{array}
\usepackage{cleveref}
\usepackage{pifont}
\usepackage{caption}
\newcommand{\cmark}{\textcolor{green!60!black}{\ding{51}}} %
\newcommand{\pmark}{\textcolor{orange!50!yellow!90!black}{\ding{51}}} %

\title{EchoXFlow: A Beamspace Echocardiography Dataset for Cardiac Motion, Flow, and Function}

\author{%
  Elias Stenhede \\
  Akershus University Hospital \\
  \And
  Joanna Sulkowska \\
  Akershus University Hospital \\
  \And
  Eivind Bjørkan Orstad \\
  Akershus University Hospital \\
  \And
  Henrik Schirmer \\
  Akershus University Hospital \\
  \And
  Arian Ranjbar \\
  Akershus University Hospital \\
}

\begin{document}

\maketitle

\begin{abstract}
    We introduce EchoXFlow, a clinical echocardiography dataset with ultrasound in its native acquisition geometry. Existing public datasets offer limited opportunities to study how cardiac anatomy, myocardial motion, and blood flow relate: Doppler velocities are typically absent or fused as RGB overlays on grayscale anatomical images, and acquisitions are released only after scan conversion and lossy vendor display processing. EchoXFlow comprises \num{35377} recordings from \num{666} routine-care examinations of \num{622} patients. It preserves the timing, geometry, and modality relationships needed for physically grounded echo learning. Each recording is retained as separable modality-specific streams: temporally resolved 1D, 2D, and 3D data alongside multiple Doppler modalities, paired with a synchronized ECG. Clinical annotations range from guideline-defined measurements to dense 2D myocardial contours and 3D left-ventricular endocardial meshes. With its associated open-source tooling, EchoXFlow enables cross-modal, acquisition-aware learning tasks that cannot be formulated from conventional scan-converted videos alone, and serves as a testbed for 4D vision and physically grounded multi-modal learning more broadly. The dataset is available at \url{https://huggingface.co/datasets/Ahus-AIM/EchoXFlow} and the tooling at \url{https://github.com/Ahus-AIM/EchoXFlow}.
\end{abstract}

\section{Introduction}
Echocardiography is the most widely used imaging modality for cardiac assessment due to its portability, safety, and ability to provide real-time, clinically useful information~\cite{paul_echocardiography_2014}. It plays a central role in diagnosing and monitoring cardiovascular diseases, which remain the leading cause of mortality worldwide~\cite{gbd_2019_diseases_and_injuries_collaborators_global_2020}. 

An echocardiographic examination is a temporally coordinated sequence of recordings across acquisition modes (hereafter called modalities) that provide complementary information about cardiac structure, function, and blood flow. While brightness-mode (B-mode) imaging supports anatomical assessment, chamber sizing, and analysis of myocardial deformation, many routine measurements require additional modalities. Doppler imaging measures the velocity of blood and tissue and is used to assess valve function, estimate intracardiac pressures, and quantify myocardial motion, with measurements interpreted relative to cardiac-cycle timing from the ECG~\cite{lang_recommendations_2015,mitchell_guidelines_2019,galderisi_standardization_2017}.
 
\begin{figure}[htpb]
    \centering
    \includegraphics[width=\linewidth]{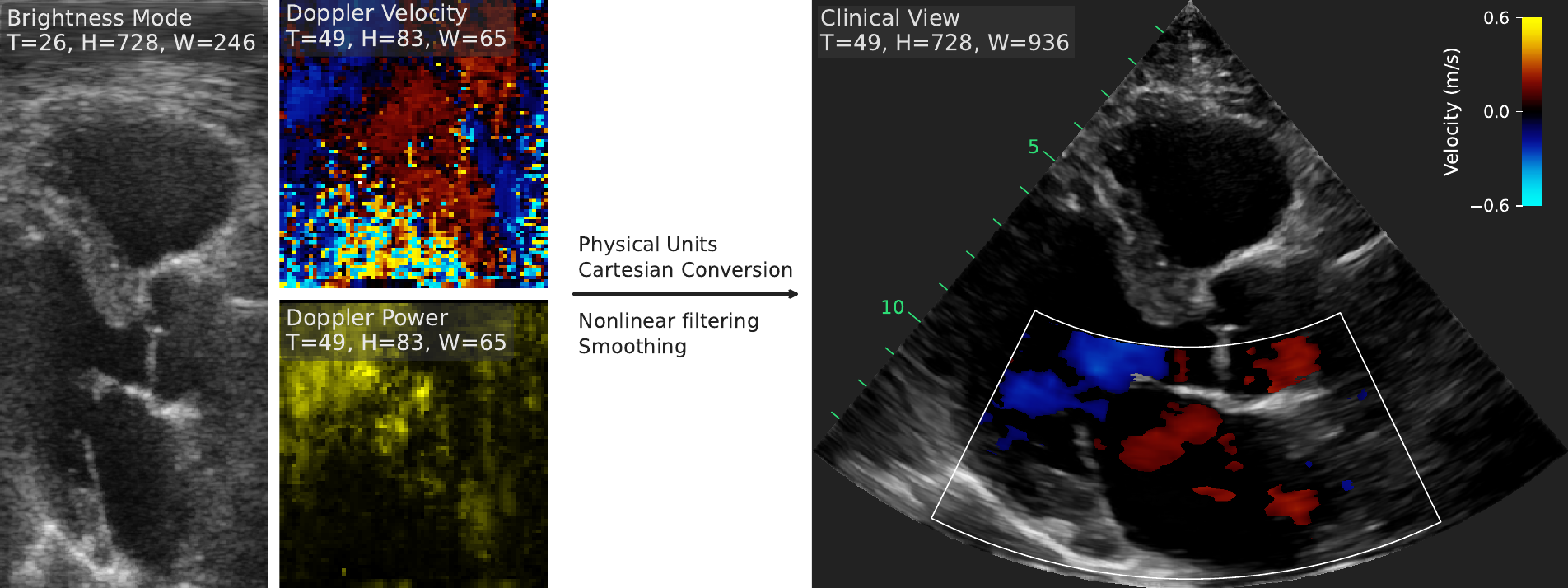}
    \caption{A frame of a color Doppler video in EchoXFlow, which can be scan-converted to a clinical view using metadata such as sector angles and depths. In clinical visualization, Doppler velocities are often smoothed and gated by combining the amplitude of the Doppler signal and the corresponding B-mode frame. Pre-scan-converted data are more efficient to store and process, and contain more information. Here, the Doppler and B-mode frames occur in a temporal 2:1 ratio.}
    \label{fig:data-conversion-color-doppler}
\end{figure}

Machine learning on echocardiography has advanced rapidly, with work spanning acquisition guidance~\cite{pasdeloup_real-time_2023}, view classification~\cite{ostvik_real-time_2019,madani_fast_2018}, segmentation~\cite{leclerc_deep_2019,ouyang_video-based_2020}, disease detection~\cite{raissi-dehkordi_contemporary_2025}, and, more recently, foundation models that solve multiple of the aforementioned tasks at once~\cite{amadou_echoapex_2024, christensen_visionlanguage_2024, kim_echofm_2025, munim_echojepa_2026, vukadinovic_comprehensive_2026}.
However, this progress has been built either on private data or on public scan-converted data, in which Doppler modalities are absent or reduced to RGB overlays within B-mode images, where vendor-specific display processing has already been applied. What reaches the model is therefore a rendering intended for human display instead of the underlying per-modality acquisition signals. \Cref{fig:data-conversion-color-doppler} illustrates the difference between pre- and post-scan-conversion.

To address these limitations, we introduce EchoXFlow, an echocardiography resource built from routine-care examinations that preserves the acquisition data. The data are released as separate streams in their probe-centered geometry, and each recording is paired with a simultaneous ECG trace which provides a shared cardiac-phase reference across recordings. Together with linked clinical annotations and tooling for working with the data, EchoXFlow makes it possible to study cardiac structure, motion, and flow. Our main contributions are:
\begin{enumerate}
    \item \textbf{EchoXFlow.} An open echocardiography dataset comprising 666 examinations containing temporally resolved 1D, 2D, and 3D B-mode data together with Doppler acquisition modes, each paired with a synchronized ECG. Acquisition geometry, timing, physical scale, and linked clinical annotations are also included.

    \item \textbf{Tooling.} Open-source code for loading beamspace arrays, mapping samples to physical coordinates, and preparing AI-model-ready inputs and targets.
\end{enumerate}

These contributions enable evaluative claims that existing public echocardiography datasets cannot support, exemplified as benchmark tasks in \Cref{sec:benchmarks}:
\begin{enumerate}
    \item \textbf{Cross-modal estimation of Doppler signals from co-acquired B-mode.} Whether a model with access only to 2D B-mode can recover the co-acquired Doppler signals: myocardial tissue velocity (Task~1) and blood-flow velocity and its local variation (Task~2).

    \item \textbf{Effect of input representation.} Whether a segmentation model benefits from training on beamspace data versus the Cartesian view (Task~3). We also assess this on Tasks 1 and 2. 

    \item \textbf{ECG-conditioned Frame-to-Video Generation.} Whether a generative model can synthesize video from a single B-mode frame, conditioned on the co-acquired ECG (Task~4).
\end{enumerate}
Beyond these, EchoXFlow supports cross-modal representation learning through ECG-based cardiac-phase alignment across non-simultaneous recordings. It also provides time-resolved 3D left-ventricular surface meshes, enabling full-cycle volumetric evaluation. More broadly, the dataset contributes to the wider ML community a rare source of densely annotated 4D data with physically coupled modalities, relevant to 4D vision and video understanding beyond cardiology.

\section{Related Work}
We review public echocardiography datasets along four axes: coverage of Doppler data, availability of 1D, 2D, and 3D acquisitions, extent of lossy post-processing applied, and the available annotations. An overview of relevant public echocardiography datasets is presented in \Cref{tab:dataset-comparison}.

\begin{table}[ht]
\centering
\caption{
Comparison of public echocardiography datasets and EchoXFlow. 
}
\footnotesize
\setlength{\tabcolsep}{0.9pt}
\renewcommand{\arraystretch}{1.1}
\begin{tabular}{
p{2.2cm}
>{\raggedleft\arraybackslash}p{1.1cm}
>{\centering\arraybackslash}p{0.7cm}
>{\centering\arraybackslash}p{0.7cm}
>{\centering\arraybackslash}p{0.7cm}
>{\centering\arraybackslash}p{0.9cm}
>{\centering\arraybackslash}p{0.95cm}
>{\centering\arraybackslash}p{0.9cm}
>{\centering\arraybackslash}p{0.9cm}
>{\centering\arraybackslash}p{2.1cm}
>{\centering\arraybackslash}p{2.1cm}
>{\centering\arraybackslash}p{2.1cm}
}
\toprule

& \multicolumn{1}{c}{\textbf{\#\,Exams}}
& \multicolumn{3}{c}{\textbf{B-mode}}
& \textbf{Video}
& \textbf{Doppler}
& \textbf{ECG}
& \textbf{Pre-SC}
& \multicolumn{2}{c}{\textbf{Annotations}} \\
\cmidrule(lr){3-5}
\cmidrule(lr){10-11}
& & \textbf{1D} & \textbf{2D} & \textbf{3D} & & & & & \textbf{Dense} & \textbf{Scalar} \\
\midrule

EchoNet-LVH
& \num{12000}
& -- & \cmark & -- & \cmark & -- & -- & --
& -- & IVS, LVID, LVPW \\

EchoNet-Dynamic
& \num{10030}
& -- & \cmark & -- & \cmark & -- & -- & --
& LV & -- \\

MIMIC-IV-Echo
& \num{7243}
& \cmark & \cmark & -- & \cmark & \pmark & \pmark & --
& -- & Multimodal \\

TMED-2
& \num{6790}
& -- & \cmark & -- & -- & -- & -- & --
& -- & View labels; AS \\

CAMUS
& \num{500}
& -- & \cmark & -- & \cmark & -- & -- & --
& LV, LA & -- \\

CardiacUDA
& $\sim$\num{100}
& -- & \cmark & -- & \cmark & -- & \pmark & --
& LV, LA, RV, RA & -- \\

TED
& \num{98}
& -- & \cmark & -- & \cmark & -- & -- & --
& LV & -- \\

MITEA
& \num{268}
& -- & -- & \cmark & -- & -- & -- & --
& LV & -- \\

CETUS
& \num{45}
& -- & -- & \cmark & -- & -- & -- & --
& LV & -- \\

STACOM~2011
& \num{15}
& -- & -- & \cmark & \cmark & -- & -- & --
& -- & -- \\

MVSeg
& \num{15}
& -- & -- & \cmark & -- & -- & -- & --
& MV & -- \\

EchoNet3D
& \num{4}
& -- & -- & \cmark & \cmark & -- & -- & \cmark
& -- & -- \\

CUBDL
& \num{2}
& -- & \cmark & -- & -- & -- & -- & \cmark
& -- & -- \\

\textbf{EchoXFlow}
& \textbf{\num{666}}
& \cmark & \cmark & \cmark & \cmark & \cmark & \cmark & \cmark
& See \Cref{tab:modality-stats} & See \Cref{tab:modality-stats} \\

\bottomrule
\end{tabular}
\par\vspace{2pt}
\begin{minipage}{\linewidth}
\footnotesize
\textit{Abbreviations.} AS: Aortic stenosis; IVS: interventricular septum thickness; LA: left atrium; LV: left ventricle; LVH: left ventricular hypertrophy; LVID: left ventricular internal dimension; LVPW: left ventricular posterior wall thickness; MV: mitral valve; Pre-SC: pre-scan-conversion (beamspace or raw sensor); \pmark: Present only as rendered overlay, not as separable signal.
\end{minipage}
\label{tab:dataset-comparison}
\end{table}

Public echocardiography datasets are predominantly scan-converted 2D B-mode recordings in standard cardiac views.  CAMUS~\cite{leclerc_deep_2019} and EchoNet-Dynamic~\cite{ouyang_video-based_2020} comprise videos with left ventricular (LV) tracings at end systole (ES) and end diastole (ED), enabling volume and ejection fraction estimation; TED~\cite{painchaud_echocardiography_2022} extends a subset of CAMUS to full-cycle LV segmentation. EchoNet-LVH~\cite{duffy_high-throughput_2022} provides videos annotated with hypertrophy-related metrics. TMED-2~\cite{huang_tmed_2022} supports view classification and aortic stenosis grading. CardiacUDA~\cite{yang_graphecho_2023} provides multi-chamber segmentation.

More recent efforts, such as MIMIC-IV-Echo~\cite{gow_mimic-iv-echo_2026}, broaden the clinical context through complete examinations and linked comprehensive clinical data, but, like other public resources, they release fused recordings intended for human display. Overall, Doppler signals are rarely released; when included at all, they are typically embedded as color overlays within B-mode images.

3D echocardiography enables volumetric assessment of cardiac anatomy and function beyond selected 2D planes. It improves chamber quantification by reducing geometric assumptions, and supports valve assessment, although spatial and temporal resolution remain practical limitations~\cite{lang_3-dimensional_2018}. Public 3D echocardiography datasets remain limited in scale and temporal coverage: STACOM 2011~\cite{tobon-gomez_benchmarking_2013} provides paired 3D ultrasound and MR data for motion tracking; CETUS~\cite{bernard_challenge_2014} and MITEA~\cite{zhao_mitea_2022} include volumes at ES/ED; MVSeg~\cite{carnahan_interactive-automatic_2019} provides full-cycle data, but only for the mitral valve. EchoNet3D~\cite{vukadinovic_automated_2025} releases a small set of full 3D acquisitions.

Public access to ultrasound data prior to scan conversion has primarily focused on beamforming and signal quality. PICMUS~\cite{liebgott_plane-wave_2016} and CUBDL~\cite{hyun_deep_2021} release raw channel (RF) data from phantoms and a small number of healthy volunteers. To our knowledge, EchoNet3D is the only public release of beamspace echocardiography data, containing 3D acquisitions from four healthy volunteers.

\section{EchoXFlow}
EchoXFlow was collected during 2025 and 2026 at the Department of Cardiology at Akershus University Hospital, which serves approximately 10\% of the Norwegian population. It consists of echocardiographic examinations of hospitalized patients, performed by experienced ultrasound operators. These patients were either admitted to the cardiac ward or referred internally from other departments to the echocardiography laboratory for evaluation of suspected cardiac disease. All data were acquired as part of standard clinical imaging workflows using a GE Vivid E95 Ultra Edition equipped with a 4Vc 4D Volume Phased Array transducer, and annotations were created using EchoPAC version 206.

Echocardiographic examinations were exported from the clinical picture archiving and communication system as DICOM files. While DICOM standardizes core metadata~\cite{national_electrical_manufacturers_association_dicom_2026}, scanner-native acquisition information may fall outside standard fields and is encoded in data elements that standard DICOM viewers do not render. In our data, these elements contain beamspace arrays, Doppler streams, ECG signals, and the metadata needed to interpret them, such as timing and geometry. We therefore developed parsers and visualizers. We validated the extraction by visual comparison with the corresponding clinical system views. Manual and semi-manual annotations were also extracted and are included with the examinations. These were generated by the ultrasound operators as part of routine patient care, following institutional workflows based on European echocardiographic guidelines~\cite{lang_recommendations_2015, galderisi_standardization_2017}.

\subsection{Data Representation and Acquisition Modes}
\begin{figure}[ht]
    \centering
    \includegraphics[width=\linewidth]{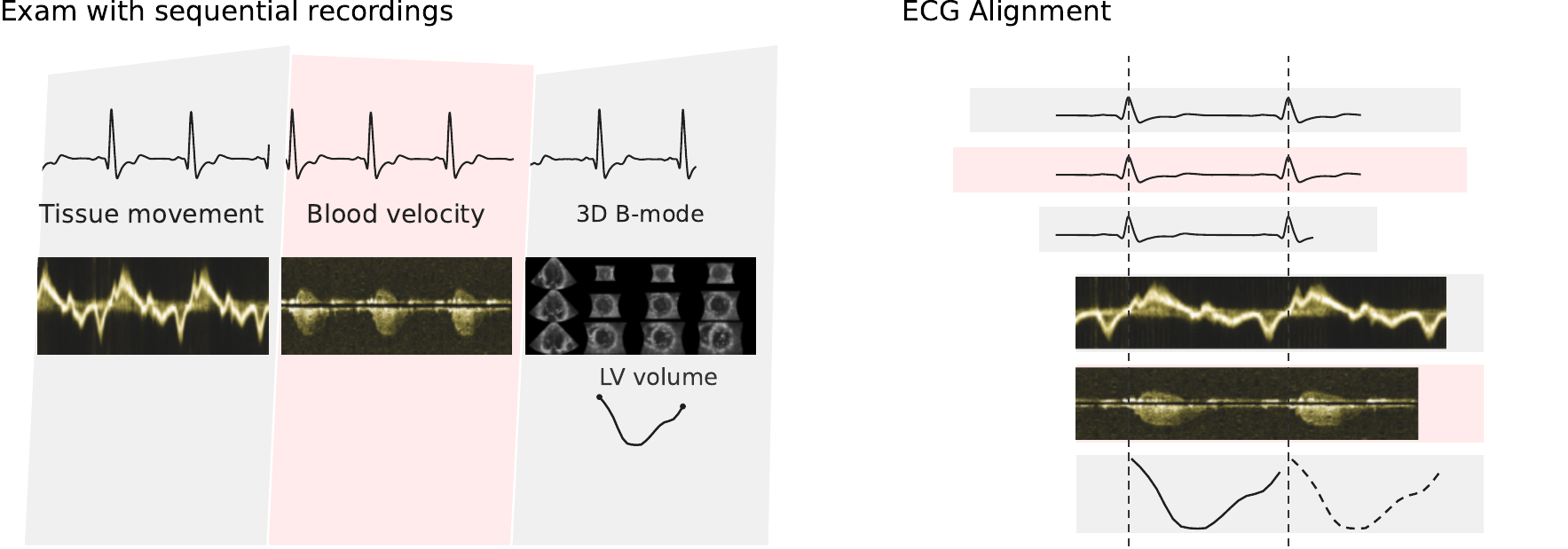}
    \caption{Recordings are paired with ECGs, enabling alignment between recordings within an exam.}
    \label{fig:r-peak-alignment}
\end{figure}

An echocardiographic examination integrates multiple acquisition modes to characterize functional, structural, and hemodynamic properties of the heart. Each examination consists of multiple short recordings that were acquired sequentially, each with an ECG trace, illustrated in~\Cref{fig:r-peak-alignment}.

Sector width, line density, imaging depth, and temporal resolution have to be traded against a finite acoustic pulse budget~\cite{ng_resolution_2011}. Thus, it is common to record both highly time-resolved 1D Doppler and B-mode data, together with 2D B-mode and Doppler data for anatomical assessment, and 3D B-mode data, which provides superior volumetric information, but at the cost of lower resolution and frame rate. In volumetric imaging, multi-beat stitching using ECG triggering may be used to increase spatial resolution, temporal resolution, or field of view by aggregating sub-volumes across cardiac cycles, at the cost of sensitivity to motion and rhythm variability. An overview of modalities in EchoXFlow is given in \Cref{tab:visual_modalities}.

\begin{figure}[t]
    \centering
    \includegraphics[width=\linewidth]{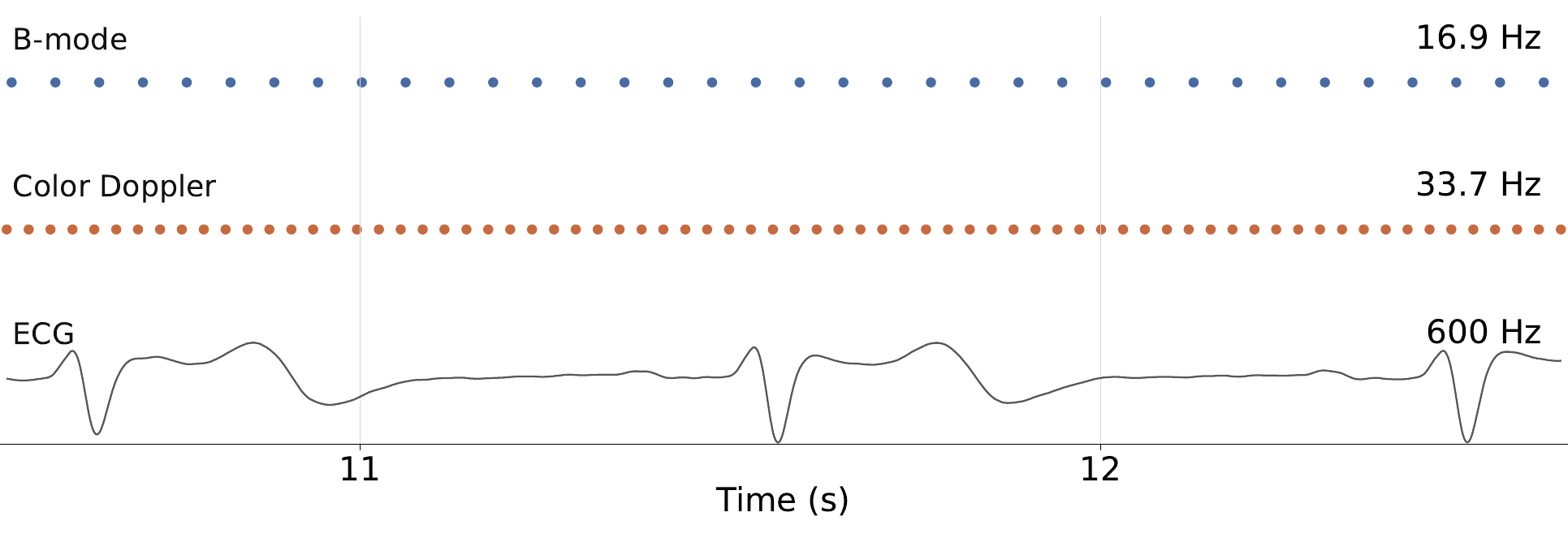}
    \caption{Structure of multimodal echocardiography recordings showing one example of the acquisition budget here prioritizing color Doppler. Modalities within a recording may have differing sampling rates; here, a factor of 1:2 between B-mode and color Doppler frames. The ECG provides a shared cardiac-phase reference for beat identification and alignment even across recordings.}
    \label{fig:temporal-alignment-frames-with-ecg}
\end{figure}

\begin{table}[ht]
    \centering
    \caption{
    Main data types in EchoXFlow, with examples showing how they are typically displayed in clinical settings. More clinical details about measurements are listed in the supplementary material.
    }
    \label{tab:visual_modalities}
    \small
    \setlength{\tabcolsep}{4pt}
    \begin{tabular}{>{\raggedright\arraybackslash}m{2.3cm}
    >{\centering\arraybackslash}m{3cm}
    >{\raggedright\arraybackslash}m{3.0cm}
    >{\raggedright\arraybackslash}m{4.5cm}
    }
    \toprule
    \textbf{Data type} & \textbf{Example view} & \textbf{Signal captured} & \textbf{Example use cases} \\
    \midrule
    \rowcolor{gray!12} 1D B-mode (M-mode) & \includegraphics[width=3cm]{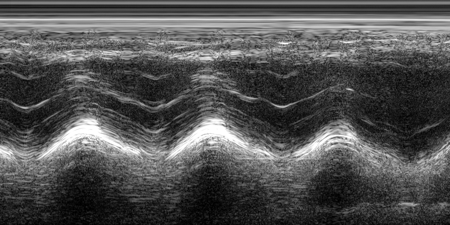} & Reflectivity along a line & Precise timing of wall and valve motion; chamber-wall thickness \\
    \rowcolor{gray!12} 2D B-mode & \includegraphics[width=3cm]{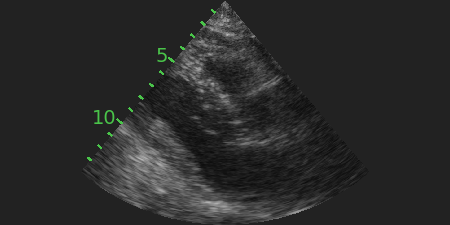} & Reflectivity in a plane & Anatomy, chamber size, and regional wall motion \\
    \rowcolor{gray!12} 3D B-mode & \includegraphics[width=3cm]{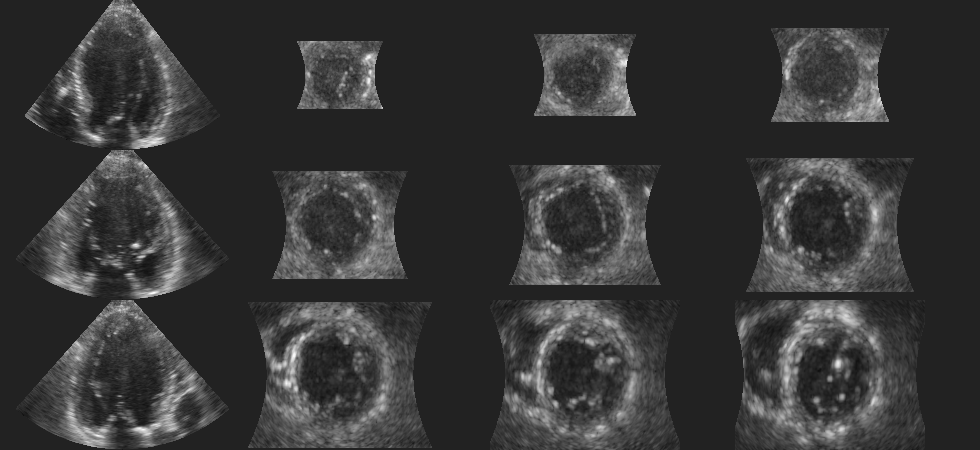} & Reflectivity in a volume & Volumetric chamber quantification and ejection fraction \\
    \rowcolor{gray!12} 1D Tissue Doppler & \includegraphics[width=3cm]{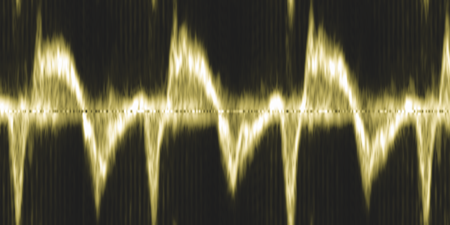} & Localized velocity & Diastolic function and peak myocardial velocities (\(s'\), \(e'\), \(a'\)) \\
    \rowcolor{gray!12} 2D Tissue Doppler & \includegraphics[width=3cm]{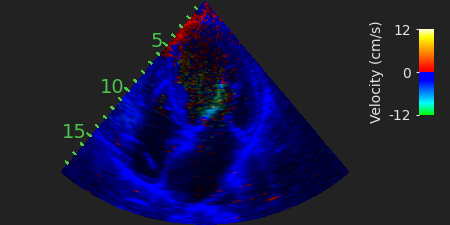} & Velocity in a plane & Regional myocardial velocity and dyssynchrony assessment \\
    \rowcolor{red!8} 1D Pulsed wave Doppler & \includegraphics[width=3cm]{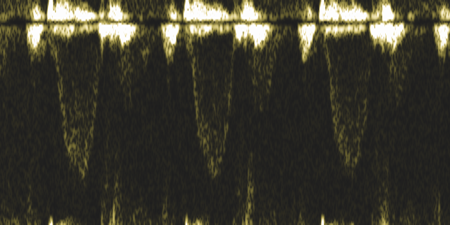} & Localized velocity & Low-velocity flow at a chosen site, e.g., mitral inflow  (\(E\), \(A\)) \\
    \rowcolor{red!8} 1D Continuous wave Doppler & \includegraphics[width=3cm]{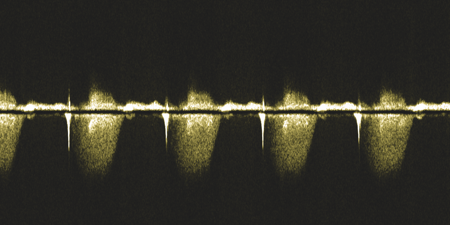} & Velocity along a line & Estimation of pressure gradients and flow over valves and defects \\
    \rowcolor{red!8} Color Doppler & \includegraphics[width=3cm]{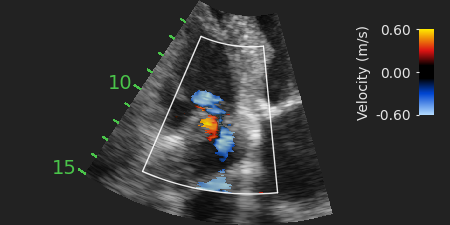} & Velocity in a plane & Screening for leakage and shunt jets \\
    \bottomrule
    \end{tabular}
    \par\vspace{2pt}
    \begin{minipage}{\linewidth}
    \footnotesize
    {\footnotesize\colorbox{gray!12}{\,Gray\,}\,modalities are used to depict the myocardium and tissue, while\,\colorbox{red!8}{\,red\,}\,modalities illustrate flowing blood.}
    \end{minipage}
\end{table}

It is also common to acquire multiple modalities concurrently, partitioning the acoustic budget between them, and acquiring interleaved temporal sequences, as illustrated in \Cref{fig:temporal-alignment-frames-with-ecg}. Using this method, B-mode data can be used to contextualize the other modalities, such as blood flow information, see \Cref{fig:data-conversion-color-doppler}. This also holds true for 1D Doppler data, where the B-mode image is used to confirm that the region of interest is placed correctly in relation to the cardiac anatomy.

Ultrasound data fall into the broad categories of 1D, 2D, or 3D B-mode or Doppler. We subdivide Doppler along two axes: blood versus tissue velocity, and for blood, pulsed- versus continuous-wave acquisition, since they have distinct clinical roles. The blood/tissue distinction reflects different filter and pulse-repetition settings: blood produces weaker backscattered ultrasound signals than myocardial tissue, so blood-flow Doppler uses a high-pass wall filter to suppress strong tissue echoes and a high pulse-repetition frequency that raises the Nyquist velocity limit for fast flow. Tissue Doppler, targeting the stronger but slower myocardial movement, instead uses a low-pass filter and a lower Nyquist limit. The pulsed/continuous choice governs spatial localization: pulsed-wave Doppler samples a user-placed volume to resolve flow in depth, but waiting for echoes between transmissions imposes a Nyquist aliasing limit, whereas continuous-wave Doppler transmits and receives simultaneously, capturing the highest-velocity jets along the entire beam with less aliasing, at the cost of depth specificity. 

\subsection{Annotations}
EchoXFlow contains both sparse clinical annotations and dense time-resolved anatomical annotations, all of which are created or adjusted by the operator performing the examination. The sparse set consists of anatomical landmarks, length measurements, traced flow envelopes, as well as spectral Doppler measurements and their corresponding sample-volume locations. Although sparse, these annotations are central to routine echocardiographic interpretation: many diagnostic variables are derived from distances, peak velocities, Doppler envelope tracings, and timing relationships. Contemporary guidelines describe in detail what these scalars are used for in clinical practice~\cite{mitchell_guidelines_2019}. The Doppler sample-volume locations are themselves meaningful as annotations; they are guideline-defined and placed at anatomical landmarks. A Doppler spectrum and its derived measurements are only interpretable given the site at which they were acquired.

The dense annotations cover 2D and 3D cardiac structures. In 2D, speckle-tracking yields time-resolved myocardial contours over a cardiac cycle; the contours act as per-frame myocardial segmentations and form the basis for segmental or, when considered jointly, global strain estimates. In 3D, left-ventricular endocardial annotations are represented as dynamic surface meshes aligned with the volumetric acquisitions, including both single-cycle and ECG-gated multi-beat stitched recordings, and can be used directly for geometric analysis or converted to alternative representations such as voxel masks or volume--time curves. Illustrations of the dense annotations and further explanations can be found in the supplementary material.

\subsection{Dataset Statistics}
EchoXFlow comprises 666 echocardiographic examinations from 622 patients. Approximately 38\% of patients were female, and the mean age was 69 years, with a standard deviation of 15 years. The dataset spans \num{35377} unique recordings, with a median of 52 recordings per examination. Per-modality counts and acquisition characteristics are summarised in \Cref{tab:modality-stats}. Every recording includes a co-registered ECG trace for cardiac-cycle timing and one or more 1D, 2D, or 3D B-mode and Doppler acquisitions; \num{145} of the \num{716} 3D B-mode recordings are themselves stitched across 2--6 cardiac beats.
Because they are time-resolved, dense annotations cover \num{63850} 2D frames and \num{5039} 3D volumes. In 2D, the majority are left-ventricular contours in apical views (A2C, A3C, A4C), with smaller subsets on the right ventricle (\num{1542} frames) and the left atrium (\num{1621} frames); in 3D, they consist of left-ventricular endocardial meshes. Sparse annotations comprise sample-volume locations for \num{5445} of the \num{5452} pulsed-wave and \num{2760} of the \num{2807} continuous-wave Doppler acquisitions, together with \num{3843} point or line annotations spanning standard echocardiographic measurements.

\begin{table}[ht]
\centering
\caption{Per-modality statistics, reported as medians and interquartile ranges.}
\footnotesize
\sisetup{
  group-separator={\,},
  group-minimum-digits=4
}
\setlength{\tabcolsep}{3pt}
\begin{tabular*}{\linewidth}{@{\extracolsep{\fill}}lrr@{}c@{}rrr@{}c@{}rrr@{}c@{}rrr@{}}
\toprule
{}
& \multicolumn{1}{c}{\textbf{Items}}
& \multicolumn{1}{c}{\textbf{Exams}}
& {}
& \multicolumn{3}{c}{\textbf{\begin{tabular}[b]{@{}c@{}}Bytes / item\\(MiB)\end{tabular}}}
& {}
& \multicolumn{3}{c}{\textbf{\begin{tabular}[b]{@{}c@{}}Sampling rate\\(Hz)\end{tabular}}}
& {}
& \multicolumn{3}{c}{\textbf{\begin{tabular}[b]{@{}c@{}}Duration/item\\(s)\end{tabular}}} \\
\cmidrule(lr){5-7} \cmidrule(lr){9-11} \cmidrule(lr){13-15}
& & {} & {}
& \multicolumn{1}{r}{\textbf{P25}} & \multicolumn{1}{r}{\textbf{P50}} & \multicolumn{1}{r}{\textbf{P75}}
& {}
& \multicolumn{1}{r}{\textbf{P25}} & \multicolumn{1}{r}{\textbf{P50}} & \multicolumn{1}{r}{\textbf{P75}}
& {}
& \multicolumn{1}{r}{\textbf{P25}} & \multicolumn{1}{r}{\textbf{P50}} & \multicolumn{1}{r}{\textbf{P75}} \\
\midrule

\textbf{Electrocardiogram} & \num{35377} & 666 & {} & $<$1 & $<$1 & $<$1 & {} & 600 & 600 & 600 & {} & 2.1 & 2.9 & 3.5 \\
\textbf{Brightness mode} & {} & {} & {} & {} & {} & {} & {} & {} & {} & {} & {} & {} & {} & {} \\
\quad 1D (M-mode) & \num{2527} & 643 & {} & 4 & 5 & 6 & {} & 496 & 496 & 496 & {} & 3.0 & 4.0 & 4.0 \\
\quad 2D (single plane) & \num{26356} & 666 & {} & 5 & 11 & 20 & {} & 12 & 35 & 49 & {} & 2.0 & 2.6 & 3.2 \\
\quad 2D (dual plane) & \num{657} & 497 & {} & 15 & 19 & 24 & {} & 23 & 24 & 28 & {} & 1.9 & 2.5 & 2.9 \\
\quad 2D (triplane) & \num{573} & 479 & {} & 40 & 49 & 62 & {} & 42 & 47 & 53 & {} & 1.8 & 2.4 & 2.9 \\
\quad 3D\textsuperscript{1} & \num{716} & 487 & {} & 124 & 182 & 240 & {} & 23 & 28 & 37 & {} & 1.9 & 2.6 & 3.3 \\
\textbf{Doppler} & {} & {} & {} & {} & {} & {} & {} & {} & {} & {} & {} & {} & {} & {} \\
\quad 2D Tissue & \num{1194} & 469 & {} & 10 & 12 & 15 & {} & 26 & 30 & 35 & {} & 3.0 & 3.0 & 4.0 \\
\quad 2D Color & \num{8451} & 663 & {} & 3 & 4 & 6 & {} & 17 & 21 & 25 & {} & 1.8 & 2.5 & 3.0 \\
\quad 1D Pulsed Wave & \num{5452} & 661 & {} & 1 & 1 & 1 & {} & 479 & 534 & 643 & {} & 3.0 & 3.8 & 4.0 \\
\quad 1D Continuous Wave & \num{2807} & 657 & {} & 1 & 2 & 2 & {} & 501 & 501 & 506 & {} & 3.0 & 4.0 & 4.0 \\
\textbf{Dense annotations} & {} & {} & {} & {} & {} & {} & {} & {} & {} & {} & {} & {} & {} & {} \\
\quad 3D segmentation & {} & {} & {} & {} & {} & {} & {} & {} & {} & {} & {} & {} & {} & {} \\
\qquad Left ventricle & \num{201} & 199 & {} & {} & {} & {} & {} & 25 & 27 & 30 & {} & 0.8 & 0.8 & 1.0 \\
\quad 2D strain/segmentation & {} & {} & {} & {} & {} & {} & {} & {} & {} & {} & {} & {} & {} & {} \\
\qquad Left ventricle, A2C & \num{443} & 334 & {} & {} & {} & {} & {} & 48 & 49 & 56 & {} & 0.9 & 1.0 & 1.1 \\
\qquad Left ventricle, A3C & \num{277} & 273 & {} & {} & {} & {} & {} & 48 & 49 & 56 & {} & 0.9 & 1.0 & 1.1 \\
\qquad Left ventricle, A4C & \num{443} & 334 & {} & {} & {} & {} & {} & 49 & 49 & 56 & {} & 0.9 & 1.0 & 1.1 \\
\qquad Left atrium, A2C & \num{9} & 9 & {} & {} & {} & {} & {} & 49 & 56 & 79 & {} & 0.9 & 0.9 & 1.1 \\
\qquad Left atrium, A4C & \num{11} & 11 & {} & {} & {} & {} & {} & 53 & 79 & 126 & {} & 0.9 & 1.0 & 1.1 \\
\qquad Right ventricle, A4C & \num{25} & 25 & {} & {} & {} & {} & {} & 49 & 49 & 60 & {} & 0.9 & 1.1 & 1.3 \\
\textbf{Sparse annotations} & {} & {} & {} & {} & {} & {} & {} & {} & {} & {} & {} & {} & {} & {} \\
\quad Sample volumes\textsuperscript{2} & {} & {} & {} & {} & {} & {} & {} & {} & {} & {} & {} & {} & {} & {} \\
\qquad 1D Pulsed Wave & \num{5445} & 661 & {} & {} & {} & {} & {} & {} & {} & {} & {} & {} & {} & {} \\
\qquad 1D Continuous Wave & \num{2760} & 652 & {} & {} & {} & {} & {} & {} & {} & {} & {} & {} & {} & {} \\
\quad Other annotations & {} & {} & {} & {} & {} & {} & {} & {} & {} & {} & {} & {} & {} & {} \\
\qquad Points and markers\textsuperscript{3} & \num{3843} & 561 & {} & {} & {} & {} & {} & {} & {} & {} & {} & {} & {} & {} \\

\bottomrule
\end{tabular*}
\par\vspace{2pt}
\begin{minipage}{\linewidth}
\footnotesize
\textit{Abbreviations.} A2C: Apical Two-Chamber View; A3C: Apical Three-Chamber View; A4C: Apical Four-Chamber View.\\
\textsuperscript{1} 571 single-beat/unstitched; 14 over 2 beats; 74 over 3 beats; 53 over 4 beats; 4 over 6 beats.\\
\textsuperscript{2} Sample volumes denote guideline-based marker locations for the 1D Doppler acquisitions.\\
\textsuperscript{3} Most common: A (1157); MV E’ septal (883); LVOT trace (744); MV E’ lateral (711); TR Vmax (551); AV trace (517).\\
\end{minipage}
\label{tab:modality-stats}
\end{table}

\section{Benchmark Tasks}\label{sec:benchmarks}
We define a set of experimental tasks designed to isolate the impact of data representation, modality, and physical consistency in echocardiographic learning, each also instantiating a general ML problem: Tasks~1 and~2 as cross-modal video-to-video translation under physical constraints and missing information (out-of-plane motion), Task~3 as segmentation under non-Euclidean input geometry, Task~4 as video generation conditioned on image and time series data.
For each of the $N$ training cases, indexed by $i$, we let \(B_i \in \mathbb{R}^{T_i \times H_i \times W_i}\) denote the B-mode sequence, and $f_\theta$ the model. In Task~1 and Task~2, the input and target frames are interleaved temporally, as shown in \Cref{fig:temporal-alignment-frames-with-ecg}.

\paragraph{Task 1: Cross-Modal Estimation of Tissue Velocity.}
We want to infer 2D tissue Doppler velocities from B-mode cine data: \(\hat{V}_i = f_{\theta}(B_i)\). With case-specific Nyquist velocity \(\nu_i > 0\), parameters are estimated by minimizing a Nyquist-normalized periodic \(L_1\) loss that treats values differing by integer multiples of \(2\nu_i\) as alias-equivalent:
\begin{equation}\label{eq:l1}
    \mathcal{L}_\text{T1}(\theta)
    =
    \frac{1}{N}\sum_{i=1}^{N}
    \text{mean}\!\left(d_{\mathrm{alias}}(\hat{V}_i,V_i;\nu_i)\right),
    \quad
    d_{\mathrm{alias}}(a,b;\nu_i)
    =
    \frac{1}{\nu_i}
    \min_{k\in\mathbb{Z}}
    \left|a-b+2k\nu_i\right|,
\end{equation}
applied elementwise and averaged over all spatiotemporal pixels inside the sector. Velocity is measured in \si{\metre/\second}.

\paragraph{Task 2: Cross-Modal Estimation of Blood Velocity.}
Unlike the myocardial tissue of Task~1, blood flow has sharp spatial velocity gradients; blood-flow signals are expected mainly in low-intensity B-mode regions, and velocity estimates at low Doppler power are more susceptible to noise (see \Cref{fig:data-conversion-color-doppler}). We want to infer blood flow velocity $V_i$, Doppler power $P_i$, and the local velocity variation $S_i$ from B-mode data: \((\hat{V}_i,\hat{P}_i,\hat{S}_i) = f_{\theta}(B_i) \in \mathbb{R}^{3 \times 2T_i \times H_i \times W_i}\), at the Doppler frame rate, which is twice the B-mode frame rate. Here, $S_i$ is the standard deviation of \({V}_i\) over each pixel's $3\times3$ neighborhood inside the color box, with neighbor differences wrapped into the Nyquist interval so that aliased neighbors count by their phase distance. With case-specific Nyquist velocity \(\nu_i > 0\), color-Doppler box \(C_i\), and valid-velocity mask \(m_i = C_i\,\mathbf{1}\{P_i > \tau_P\}\,\mathbf{1}\{B_i < \tau_B\}\) selecting pixels with sufficient Doppler power in dark B-mode regions, we minimize a masked \(L_1\) loss:
\begin{equation}\label{eq:l2}
    \mathcal{L}_\text{T2}(\theta)
    =
    \frac{1}{N}\sum_{i=1}^{N}
    \left[
    \text{mean}\!\left(|\hat{P}_i - P_i|\right)
    +
    \text{mean}_{m_i}\!\left(|\hat{V}_i - V_i|\right)
    +
    \text{mean}_{m_i}\!\left(|\hat{S}_i - S_i|\right)
    \right],
\end{equation}
where \(\text{mean}_{m_i}\) averages only over pixels where \(m_i = 1\). In other words, velocity and variation are only supervised where the signal is reliable. 

\paragraph{Task 3: Segmentation of Left Ventricular Cavity and Myocardium.}
We predict per-pixel class probabilities \(\hat{M}_i = f_{\theta}(B_i)\) over background, LV cavity, and myocardium for the B-mode cine data, with reference masks \(M_i\) rasterized on every frame from the nearest annotated contour in time. Parameters are estimated by minimizing a soft Dice loss over the whole clip:
\begin{equation}\label{eq:l3}
    \mathcal{L}_\text{T3}(\theta)
    =
    \frac{1}{N}\sum_{i=1}^{N}
    \mathcal{L}_{\mathrm{Dice}}(\hat{M}_i, M_i).
\end{equation}

\paragraph{Task 4: ECG-Conditioned Frame-to-Video Generation.}
We aim to generate temporally coherent B-mode cine sequences from a single reference frame and a synchronized ECG signal. The reference frame provides the anatomical appearance, while the ECG provides temporal information about the cardiac cycle. We train a conditional video diffusion model to reconstruct the sequence $B_i$ from a noisy version $B_{i,t}$, conditioned on the reference frame $B_i^{(0)}$ and ECG signal $E_i$. At diffusion timestep $t$, the model predicts the velocity parameterization of the denoising process, $\hat{v}_{i,t} = f_\theta(B_{i,t},t,B_i^{(0)},E_i)$, by minimizing
\begin{equation}\label{eq:l4}
\mathcal{L}_\text{T4}(\theta)
=
\frac{1}{N}\sum_{i=1}^{N}
\mathbb{E}_{t,\epsilon}\left[
\text{mean}\!\left(|\hat{v}_{i,t}-v_{i,t}|^2\right)
\right],
\end{equation}
where $v_{i,t}$ is the diffusion velocity target. This task evaluates whether synchronized ECG information can guide the generation of anatomically consistent cardiac motion from a static ultrasound frame.

\section{Benchmark Evaluation}
All tasks were evaluated by 5-fold cross-validation on the patient-level split provided with EchoXFlow, at three resolutions ($128\times128$, $256\times256$ and $320\times320$), with every model trained once on beamspace data and once on scan-converted data. For Tasks 1--3, we trained five framewise architectures that process each frame independently and five time-aware architectures that process the whole clip, and compared them with a baseline that always predicts the average training target value at each pixel. All models were initialized from scratch without pretrained weights; all predictions were scan-converted to a common Cartesian grid before computing the metrics. Task 4 was evaluated by comparing generated and real clips in seven pretrained feature spaces.

Results for resolutions other than $256\times256$, full training hyperparameters, and architecture details are given in the supplementary material. \Cref{tab:results-r256} shows the results for Tasks 1-3. Time-aware models outperform framewise models on Tasks~1 and~2, with a large margin on velocity and power, indicating that temporal context carries the cross-modal signal, which aligns well with intuition. Comparing input representations, results are mixed. The Cartesian representation seems suitable for segmentation, yielding a higher Dice score for all models at $256\times256$ and $320\times320$ (and for all but one at $128\times128$), whereas the beamspace representation seems suitable for Doppler velocity and power, where the measured velocity directions align with the columns in the beamspace representation; for the velocity variation, several framewise models favour the scan-converted input (\Cref{tab:domain-comparison-r256} in Appendix \Cref{app:additional-results}).

The generative model in Task 4 was evaluated using seven feature extractors. Four are frame-level encoders (EchoPrime view classifier, EchoFine, EchoCLIP and Inception), scored with the Fréchet distance. Three are clip-level encoders (PanEcho, EchoJEPA and the EchoPrime MViT video encoder), scored with the kernel distance (KD). Each metric compares generated clips with the real clips of the same validation examinations and is averaged over the A2C, A3C and A4C views.                       
\Cref{tab:task4-generation-r256} compares models trained on beamspace data with models trained on scan-converted data. At $256\times256$, the beamspace model achieved lower distances in all seven feature spaces; across the three resolutions, the only exception is the EchoPrime MViT space at $128\times128$ (\Cref{tab:task4-generation-cv} in Appendix \Cref{app:additional-results}). The largest gains appeared in the clip-level spaces, which are sensitive to motion and temporal coherence. One possible explanation for the lower distances is that in beamspace the model does not need to reproduce the sector geometry, and the speckle pattern appears along the beam rather than varying with acquisition angle as it does after scan conversion. The model can therefore possibly instead devote its capacity to the anatomy and its motion.

Resampling to equal input and output sizes was used in this comparison, which masks a practical advantage of beamspace: at native resolution, beamspace tensors are smaller than their isotropically scan-converted counterparts, a difference that is particularly marked in 3D. Future work could investigate architectures that explicitly respect the polar/spherical geometry, due to the computational advantages. The Doppler reference itself contains noise and artifacts, so the irreducible error floor for Tasks~1 and~2 is unknown; the absolute $L_1$ values should be interpreted relative to each other.

\begin{table}[ht]
    \centering
    \caption{Cross-validated benchmark results for Tasks 1-3 at $256\times256$ resolution.}
    \label{tab:results-r256}
    \footnotesize
    \setlength{\tabcolsep}{2.7pt}
    \sisetup{detect-weight=true, detect-family=true}
    \begin{tabular*}{\linewidth}{@{\extracolsep{\fill}}ll!{\extracolsep{\fill}}S[table-format=1.2]@{\extracolsep{0pt}\,$\pm$\,}S[table-format=1.2]!{\extracolsep{\fill}}S[table-format=2.2]@{\extracolsep{0pt}\,$\pm$\,}S[table-format=1.2]!{\extracolsep{\fill}}S[table-format=1.3]@{\extracolsep{0pt}\,$\pm$\,}S[table-format=1.3]!{\extracolsep{\fill}}S[table-format=1.2]@{\extracolsep{0pt}\,$\pm$\,}S[table-format=1.2]!{\extracolsep{\fill}}S[table-format=2.1]@{\extracolsep{0pt}\,$\pm$\,}S[table-format=1.1]@{}}
    \toprule
    \textbf{Model} & \textbf{Input domain} & \multicolumn{2}{c}{\textbf{Task 1} $\downarrow$} & \multicolumn{6}{c}{\textbf{Task 2} $\downarrow$} & \multicolumn{2}{c}{\textbf{Task 3} $\uparrow$} \\
    \cmidrule(lr){3-4}\cmidrule(lr){5-10}\cmidrule(lr){11-12}
    & & \multicolumn{2}{c}{\textbf{Velocity}} & \multicolumn{2}{c}{\textbf{Velocity}} & \multicolumn{2}{c}{\textbf{Power}} & \multicolumn{2}{c}{\textbf{Variation}} & \multicolumn{2}{c}{\textbf{Dice (\%)}} \\
    \midrule
    \multicolumn{12}{@{}l}{\textbf{Baseline}} \\
    \quad Temporal mean & Beamspace & 2.38 & 0.12 & 16.47 & 0.26 & 0.174 & 0.006 & 3.64 & 0.08 & 50.8 & 0.6 \\
     & Scan-converted & 2.38 & 0.12 & 16.47 & 0.26 & 0.176 & 0.006 & 3.91 & 0.08 & 51.1 & 0.6 \\
    \multicolumn{12}{@{}l}{\textbf{Framewise}} \\
    \quad 2D U-Net & Beamspace & 2.21 & 0.12 & 15.68 & 0.29 & 0.153 & 0.003 & 2.42 & 0.08 & 82.8 & 0.8 \\
     & Scan-converted & 2.21 & 0.13 & 15.73 & 0.29 & 0.153 & 0.002 & 2.40 & 0.08 & 83.3 & 0.7 \\
    \addlinespace[1pt]
    \quad ConvNeXt V2 Atto & Beamspace & 2.18 & 0.10 & 15.87 & 0.25 & 0.154 & 0.002 & 2.47 & 0.07 & 83.1 & 0.7 \\
     & Scan-converted & 2.25 & 0.11 & 16.02 & 0.24 & 0.155 & 0.003 & 2.41 & 0.08 & 83.5 & 0.8 \\
    \addlinespace[1pt]
    \quad SegFormer-B0 & Beamspace & 2.18 & 0.11 & 15.93 & 0.31 & 0.153 & 0.003 & 2.75 & 0.31 & 83.6 & 0.7 \\
     & Scan-converted & 2.31 & 0.20 & 15.76 & 0.09 & 0.156 & 0.005 & 2.44 & 0.08 & 83.9 & 0.8 \\
    \addlinespace[1pt]
    \quad DINOv3 ViT-S/16 & Beamspace & 2.17 & 0.12 & 15.55 & 0.22 & 0.151 & 0.004 & 2.42 & 0.08 & 83.3 & 0.8 \\
     & Scan-converted & 2.19 & 0.10 & 15.66 & 0.17 & 0.153 & 0.003 & 2.41 & 0.05 & 83.7 & 0.8 \\
    \addlinespace[1pt]
    \quad MambaOut-Femto & Beamspace & 2.22 & 0.15 & 15.63 & 0.31 & 0.152 & 0.003 & 2.45 & 0.13 & 83.2 & 0.7 \\
     & Scan-converted & 2.26 & 0.13 & 15.62 & 0.30 & 0.153 & 0.003 & 2.41 & 0.07 & 83.9 & 0.9 \\
    \multicolumn{12}{@{}l}{\textbf{Time-aware}} \\
    \quad 3D U-Net & Beamspace & 1.24 & 0.07 & 14.18 & 0.40 & 0.126 & 0.003 & 2.37 & 0.08 & 83.1 & 0.9 \\
     & Scan-converted & 1.24 & 0.06 & 14.13 & 0.38 & 0.127 & 0.002 & 2.38 & 0.07 & 83.7 & 0.7 \\
    \addlinespace[1pt]
    \quad Video Swin-T & Beamspace & 1.38 & 0.09 & 14.56 & 0.28 & 0.133 & 0.002 & 2.39 & 0.06 & 82.9 & 0.9 \\
     & Scan-converted & 1.40 & 0.10 & 14.61 & 0.31 & 0.133 & 0.003 & 2.39 & 0.10 & 83.4 & 0.7 \\
    \addlinespace[1pt]
    \quad MViT-v2-S & Beamspace & 1.26 & 0.07 & \bfseries 13.83 & 0.36 & 0.124 & 0.002 & 2.36 & 0.07 & 83.8 & 0.7 \\
     & Scan-converted & 1.28 & 0.06 & 13.93 & 0.34 & 0.125 & 0.003 & \bfseries 2.36 & 0.09 & \bfseries 84.5 & 0.8 \\
    \addlinespace[1pt]
    \quad UniFormer-S 32f & Beamspace & 1.28 & 0.07 & 14.00 & 0.37 & 0.126 & 0.002 & 2.38 & 0.08 & 83.5 & 0.8 \\
     & Scan-converted & 1.30 & 0.07 & 14.02 & 0.37 & 0.128 & 0.003 & 2.37 & 0.07 & 84.1 & 0.8 \\
    \addlinespace[1pt]
    \quad U-Mamba-Bot & Beamspace & \bfseries 1.19 & 0.07 & 14.05 & 0.27 & \bfseries 0.122 & 0.003 & 2.37 & 0.05 & 83.3 & 0.6 \\
     & Scan-converted & 1.21 & 0.07 & 14.03 & 0.43 & 0.124 & 0.002 & 2.37 & 0.09 & 83.8 & 0.7 \\
    \bottomrule
    \end{tabular*}
    \par\vspace{1pt}
    \begin{minipage}{\linewidth}
    \footnotesize
    Mean $\pm$ standard deviation across the folds. $\downarrow$ lower is better, $\uparrow$ higher is better. Velocities and variation in \SI{}{\centi\meter/\second}.
    \end{minipage}
\end{table}

\begin{table}[ht]
    \centering
    \caption{ECG-conditioned generation quality (Task~4) at $256\times256$ resolution, by training domain.}
    \label{tab:task4-generation-r256}
    \footnotesize
    \setlength{\tabcolsep}{3pt}
    \begin{tabular*}{\linewidth}{@{\extracolsep{\fill}}lccc@{}}
    \toprule
    \textbf{Feature space} & \textbf{Beamspace} & \textbf{Scan-converted} & \textbf{Ratio} \\
    \midrule
    \multicolumn{4}{@{}l}{\textbf{Frame-level (FD $\downarrow$)}} \\
        \quad EchoCLIP & $\mathbf{0.163} \pm 0.008$ & $0.235 \pm 0.005$ & $1.44\times$ \\
    \quad EchoFine & $\mathbf{5.35} \pm 0.89$ & $7.54 \pm 1.77$ & $1.41\times$ \\
    \quad EchoPrime view & $\mathbf{0.373} \pm 0.090$ & $0.451 \pm 0.103$ & $1.21\times$ \\
    \quad Inception (FID) & $\mathbf{14.8} \pm 0.3$ & $39.8 \pm 1.3$ & $2.69\times$ \\
    \multicolumn{4}{@{}l}{\textbf{Clip-level (KD $\downarrow$)}} \\
    \quad EchoJEPA & $\mathbf{6.54} \pm 0.99$ & $27.0 \pm 0.6$ & $4.12\times$ \\
    \quad EchoPrime MViT & $\mathbf{1.76} \pm 0.13$ & $5.60 \pm 0.16$ & $3.19\times$ \\
    \quad PanEcho & $\mathbf{0.209} \pm 0.018$ & $0.234 \pm 0.015$ & $1.12\times$ \\
    \bottomrule
    \end{tabular*}
    \par\vspace{2pt}
    \begin{minipage}{\linewidth}
    \footnotesize
    Mean $\pm$ standard deviation across the folds. Ratio: scan-converted / beamspace mean (above 1 favours beamspace).
    \end{minipage}
\end{table}

\section{Limitations}
EchoXFlow inherits limitations from its acquisition setting. Doppler ultrasound measures only the beam-aligned velocity component, so estimates supervised against Doppler targets (e.g., Tasks 1 and 2) do not measure motion perpendicular to the beam. This is a well-known fact of Doppler imaging, and guidelines are formulated with this in mind~\cite{mitchell_guidelines_2019}. ECG-based cardiac-phase alignment, used for 3D beat stitching and optionally for cross-recording temporal alignment, assumes sufficiently regular rhythm and can introduce stitching artifacts under arrhythmias~\cite{lang_3-dimensional_2018}.

All recordings come from a single vendor (GE Vivid E95) at a single university hospital, so vendor-specific beam sequencing and institutional referral patterns propagate into the data. Transfer to other vendors or patient settings should not be assumed without external validation. Nonetheless, using beamspace data as opposed to scan-converted data bypasses some of the vendor-specific post-processing steps, which could be beneficial but remains to be explored.

Annotations were produced by clinical operators during routine reporting, and therefore reflect inter-operator variability and indication-driven missingness that is not at random. Reference standards are echocardiographic, and not, for example, cross-validated to cardiac magnetic resonance imaging, which is often considered the gold standard in volumetric assessment~\cite{kinno_comparison_2017}. Per Scandinavian clinical convention, race and ethnicity are never recorded; age and sex are shared in aggregated form, but these attributes are not released at the patient level to maximize anonymity. Subgroup analyses along these axes are therefore not possible. Finally, EchoXFlow is cross-sectional and does not support longitudinal analyses, nor does it contain diagnostic labels.

\section{Ethical Considerations}
Data were collected retrospectively under a waiver approved by the Norwegian Directorate of Health (case 24/45684-5, January 24, 2025), in accordance with Chapter 5, section 29 of the Norwegian Health Personnel Act. All released data are anonymized, with no patient demographics besides aggregate statistics, and no names, dates, or other direct or indirect identifiers included. The release is built by allow-listing: only decoded acquisition payloads and a fixed set of structural attributes are exported. Nonetheless, users must not attempt re-identification or link the dataset to external records that could enable it.

\section{Conclusion}
We introduced EchoXFlow, a clinical echocardiography dataset of \num{666} routine-care examinations from \num{622} patients that preserves each modality and the ECG as separate signal streams in their native acquisition geometry.  Together with open-source tooling for loading, scan conversion, and benchmark preparation, EchoXFlow makes it possible to study cardiac structure, motion, and flow as the coupled, physically grounded signals they are in clinical practice.

Our benchmark tasks serve as starting points for further study and enable evaluations that existing public datasets cannot support: recoverability of tissue and blood-flow Doppler from co-acquired B-mode, the effect of native beamspace versus Cartesian input on segmentation and video generation. Beyond these, the released streams, ECG alignment, and dense 3D supervision open a broader space of cross-modal and acquisition-aware learning problems.

Looking forward, research enabled by this dataset may have acquisition consequences. If, in the future, a modality could be reliably inferred from another, the limited acoustic budget could be reallocated to modalities that carry complementary information, increasing spatial or temporal resolution where it matters most and potentially improving diagnostic assessment. Lastly, the same data also offer a clinically grounded substrate for 4D vision, cross-modal translation, and physically consistent representation learning beyond cardiology.

EchoXFlow and associated resources are publicly available under the CC BY-NC-SA 4.0 license: the dataset at \url{https://huggingface.co/datasets/Ahus-AIM/EchoXFlow} and the code at \url{https://github.com/Ahus-AIM/EchoXFlow}.

\begin{ack}
Akershus University Hospital funded this research; the authors declare no competing interests.
\end{ack}

\newpage

\bibliographystyle{unsrtnat}
\bibliography{references}

\newpage
\appendix

\section{Training Setup}\label{app:training-setup}
This appendix is intended to give sufficient detail to reimplement the benchmarks of Tasks~1--4. All trained models share one optimizer, one random seed, and one set of patient-level cross-validation folds, in which all examinations of a patient fall in the same fold. In Tasks~1--3, ten methods share one stopping rule and one evaluation protocol; the methods differ in their backbone and decoder, and the tasks differ in their head, loss, and data filters. Task~4 trains a video diffusion model with its own architecture, objective, and evaluation. Every model is trained from random initialization on single-channel B-mode input, at three input resolutions ($128\times128$, $256\times256$, and $320\times320$) and in two input domains (beamspace and scan-converted).

\subsection{Network architectures}
\paragraph{Tasks 1-3} The benchmark contains a non-trainable baseline and ten trainable methods in two groups (\Cref{tab:methods}). Framewise methods process each frame independently, with time folded into the batch dimension. Time-aware methods process the clip as a spatiotemporal volume. Two of the methods are U-Nets~\cite{ronneberger_u-net_2015} with a shared design: six channel stages of widths 32, 64, 128, 256, 320, and 320 and five downsampling levels with spatial stride 2 and temporal stride 1, each stage a replicate-padded convolutional block with instance normalization and PReLU activations wrapped in a single residual unit, and a decoder of transposed convolutions. The 2D U-Net uses $1\!\times\!3\!\times\!3$ kernels throughout. The 3D U-Net uses $3\!\times\!3\!\times\!3$ kernels, except that the first encoder stage uses $1\!\times\!3\!\times\!3$ for Task~1 and the first three encoder stages use $1\!\times\!3\!\times\!3$ for Task~2, which delays temporal mixing where the B-mode frame rate is lower. The image backbones (ConvNeXt V2~\cite{woo_convnext_2023}, SegFormer~\cite{xie_segformer_2021}, DINOv3~\cite{simeoni_dinov3_2025}, and MambaOut~\cite{yu_mambaout_2025}) are taken from timm and segmentation\_\allowbreak models\_\allowbreak pytorch with a single input channel; ConvNeXt V2 Atto is paired with a feature pyramid network (FPN) decoder~\cite{lin_feature_2017}, SegFormer-B0 with its all-MLP decoder, and DINOv3 ViT-S/16 (blocks 9--11) and MambaOut-Femto with a light FPN-style decoder. The video backbones (Video Swin-T~\cite{liu_video_2022} and MViT-v2-S~\cite{li_mvitv2_2022} from torchvision, and a reimplementation of UniFormer-S~\cite{li_uniformer_2022}) are paired with a 3D FPN decoder of width 32 and trilinear upsampling to the full clip. U-Mamba-Bot~\cite{ma_u-mamba_2024} is a 3D residual U-Net without temporal downsampling, with a Mamba block~\cite{gu_mamba_2024} over the flattened bottleneck tokens. All inputs are replicate-padded to a multiple of the backbone stride and the outputs are cropped back.

\begin{table}[ht]
    \centering
    \caption{Benchmark methods.}
    \label{tab:methods}
    \footnotesize
    \setlength{\tabcolsep}{6pt}
    \begin{tabular*}{\linewidth}{@{\extracolsep{\fill}}llr@{}}
        \toprule
        \textbf{Model} & \textbf{Backbone and decoder} & \textbf{Parameters (millions)} \\
        \midrule
        \multicolumn{3}{@{}l}{\textbf{Baseline}} \\
        \quad Temporal mean & Per-pixel mean of the training targets & --- \\
        \multicolumn{3}{@{}l}{\textbf{Framewise}} \\
        \quad 2D U-Net & U-Net, $1\!\times\!3\!\times\!3$ kernels & 5 \\
        \quad ConvNeXt V2 Atto & ConvNeXt V2 Atto, FPN decoder & 4 \\
        \quad SegFormer-B0 & MiT-B0, all-MLP decoder & 4 \\
        \quad DINOv3 ViT-S/16 & ViT-S/16, FPN-style decoder & 22 \\
        \quad MambaOut-Femto & MambaOut-Femto, FPN-style decoder & 6 \\
        \multicolumn{3}{@{}l}{\textbf{Time-aware}} \\
        \quad 3D U-Net & U-Net, $3\!\times\!3\!\times\!3$ kernels & 15 \\
        \quad Video Swin-T & Swin3D-T, 3D FPN decoder & 30 \\
        \quad MViT-v2-S & MViT-v2-S, 3D FPN decoder & 34 \\
        \quad UniFormer-S 32f & UniFormer-S, 3D FPN decoder & 21 \\
        \quad U-Mamba-Bot & 3D residual U-Net, Mamba bottleneck & 11 \\
        \bottomrule
    \end{tabular*}
    \par\vspace{2pt}
    \begin{minipage}{\linewidth}
    \footnotesize
    \end{minipage}
\end{table}

The task heads are shared across methods (\Cref{tab:net-per-task}). For the Doppler tasks, the target is defined in physical units, while only movement in pixels per frame is observable for the network. The Task~1 output is therefore multiplied by a per-clip factor equal to the frame rate times the depth span per radial pixel, and the Task~2 velocity output is multiplied by the case-specific Nyquist velocity. In Task~1, the network predicts the velocity in each interval between consecutive B-mode frames, so a clip of $N$ B-mode frames yields $N-1$ targets. In Task~2, the head trilinearly upsamples its temporal axis by a factor of 2 to recover the Doppler frame rate.

\begin{table}[ht]
    \centering
    \caption{Per-task heads, shared by all methods.}
    \label{tab:net-per-task}
    \footnotesize
    \setlength{\tabcolsep}{6pt}
    \begin{tabular*}{\linewidth}{@{\extracolsep{\fill}}lcl@{}}
        \toprule
        \textbf{Task} & \begin{tabular}[b]{@{}r@{}}\textbf{Output channels}\end{tabular} & \textbf{Head specifics} \\
        \midrule
        Tissue Doppler & 1 & Velocity per B-mode frame interval, unit conversion \\
        Color Doppler & 3 & Velocity, power, variation; \(\times 2\) temporal; Nyquist-scaled velocity \\
        Segmentation & 3 & Final \(1\!\times\!1\!\times\!1\) projection to background, LV cavity, and myocardium \\
        \bottomrule
    \end{tabular*}
\end{table}

Beamspace inputs are the native beam grid (range by angle), bilinearly resized. Scan-converted inputs are produced by linear scan conversion onto a grid covering the bounding box of the sector, with zeros outside the sector, followed by a bilinear resize to the square resolution; beamspace and scan-converted variants therefore share parameter and compute budget. B-mode values are scaled to $[0,1]$ and min--max normalized per clip for Tasks~1 and~2 and per frame for Task~3.

\paragraph{Task 4} Given the first B-mode frame of a clip and the ECG over the clip, the Task~4 model generates the remaining 31 frames of a 32-frame clip. Clips are drawn from apical two-, three-, and four-chamber recordings, identified by the EchoPrime view classifier~\cite{vukadinovic_comprehensive_2026} with a top-1 probability of at least 0.85. The ECG conditioning for each generated frame is a \SI{1.4}{\second} window (\SI{1.2}{\second} before to \SI{0.2}{\second} after the frame). The ECG is sampled at \SI{600}{\hertz} and robustly standardized per recording and clipped; samples outside the recorded ECG are set to zero.

The model is a factorized spatiotemporal U-Net operating on $2\times2$ pixel-packed frames (\Cref{tab:task4-net}). Each residual block applies, twice, group normalization (8 groups), SiLU, and a factorized convolution ($1\!\times\!3\!\times\!3$ spatial, then $3\!\times\!1\!\times\!1$ temporal), with a $1\!\times\!1\!\times\!1$ skip projection where the width changes. The encoder has three stages of widths 128, 256, and 512 with spatial downsampling, and the decoder mirrors it with skip concatenation. The bottleneck halves the spatial resolution and applies two residual blocks. The input consists of the 31 noisy future frames, the clean first frame repeated over the 31 frames, and two coordinate channels: depth and beam angle (in radians) for beamspace input, and axial and lateral position for scan-converted input, with distances divided by \SI{20}{\centi\meter}.

Every frame receives a 512-dimensional embedding that sums the diffusion-step embedding and an ECG embedding. The ECG embedding is the output of a 1D convolutional ECG encoder for the frame's window. A linear projection of this embedding is added per channel to the features of every residual block, between its two convolutions.

\begin{table}[ht]
    \centering
    \caption{Video diffusion model.}
    \label{tab:task4-net}
    \footnotesize
    \setlength{\tabcolsep}{4pt}
    \begin{tabular*}{\linewidth}{@{\extracolsep{\fill}}ll@{}}
        \toprule
        \textbf{Component} & \textbf{Value} \\
        \midrule
        Pixel packing        & $2\times2$ space-to-channel \\
        Input channels       & 10 (noisy future frames 4, first frame 4, coordinates 2) \\
        Stage widths         & 128, 256, 512 \\
        Residual block       & $2\times$ [GroupNorm, SiLU, $1\!\times\!3\!\times\!3$ then $3\!\times\!1\!\times\!1$ convolution] \\
        Bottleneck           & $2\times$ spatial downsampling; 2 residual blocks \\
        Frame embedding      & 512-d: diffusion step $+$ ECG, added in every block \\
        ECG encoder          & 1D CNN, 16/32/64 channels, kernel size 15 \\
        Parameters (millions) & 43 \\
        \bottomrule
    \end{tabular*}
    \par\vspace{2pt}
    \begin{minipage}{\linewidth}
    \footnotesize
    \end{minipage}
\end{table}

\subsection{Optimization}
All trained models share the recipe in \Cref{tab:optim}. Tensors of rank \(>1\) (convolutional and linear weights, including the heads) are updated by the Muon optimizer~\cite{jordan_muon_2024}, with convolution kernels flattened to matrices; the remaining 1D parameters (biases and normalization parameters) are updated by AdamW. Both branches share the learning rate and use no weight decay. For models that exceed GPU memory at four clips, the effective batch of four is reached by gradient accumulation (\Cref{tab:optim}). The temporal-mean baseline is non-trainable: it predicts, for every frame, the per-pixel mean of the training targets, computed in the same coordinate space as the target; for Task~3 the logarithms of the mean class probabilities serve as constant logits.

Each epoch draws one random clip window from every training recording (Tasks~1 and~2) or annotated panel (Task~3), in an order seeded by the epoch. A fixed subset of the training examinations (12.5\%) is held out for model selection. After every epoch, the loss on this selection subset is evaluated and averaged with equal weight per examination. Training stops when this selection loss has not improved for three epochs, or after 25 epochs, and the weights of the best epoch are evaluated.

Task~4 uses the same optimizer and learning rate with a batch of one clip because of GPU memory. It trains for a fixed 10 epochs without a selection set, and generates from an exponential moving average (EMA) of the weights after the last epoch.

\begin{table}[ht]
    \centering
    \caption{Optimization recipe.}
    \label{tab:optim}
    \footnotesize
    \setlength{\tabcolsep}{4pt}
    \begin{tabular*}{\linewidth}{@{\extracolsep{\fill}}lll@{}}
        \toprule
        \textbf{Setting} & \textbf{Tasks 1--3} & \textbf{Task 4} \\
        \midrule
        Optimizer (rank \(>1\)) & \multicolumn{2}{l}{Muon, Nesterov momentum 0.95, 5 iterations} \\
        Optimizer (rank 1)      & \multicolumn{2}{l}{AdamW, \((\beta_1, \beta_2)=(0.9, 0.999)\), \(\varepsilon=10^{-8}\)} \\
        Learning rate           & \multicolumn{2}{l}{\(3.7\!\times\!10^{-3}\), constant, no warm-up} \\
        Weight decay            & \multicolumn{2}{l}{0} \\
        Gradient clipping       & \multicolumn{2}{l}{None} \\
        Precision               & \multicolumn{2}{l}{bfloat16 mixed} \\
        Random seed             & \multicolumn{2}{l}{Fixed across data sampling, initialization, and folds} \\
        Batch size              & 4\textsuperscript{1} & 1 \\
        Epochs                  & Early stopping, patience 3, at most 25 & 10 \\
        Evaluated weights       & Best epoch on 1/8 of training examinations & EMA, decay 0.995, after epoch 10 \\
        Augmentation            & None & ECG only, polarity flip, \(p=0.5\) \\
        \bottomrule
    \end{tabular*}
    \par\vspace{2pt}
    \begin{minipage}{\linewidth}
    \footnotesize
    \textsuperscript{1} Gradient accumulation for Video Swin-T, MViT-v2-S, UniFormer-S 32f and for the 2D U-Net at $320\times320$ resolution.
    \end{minipage}
\end{table}

\subsection{Losses and thresholds}
\Cref{tab:losses} summarizes the loss for each task. The Nyquist-wrapped \(L_1\) on tissue velocity is defined as in \Cref{eq:l1}: the prediction--target residual is wrapped into \([-\nu_i, \nu_i)\) using the case-specific Nyquist velocity \(\nu_i\) and divided by \(\nu_i\), averaged over the pixels inside the sector. The Color Doppler loss in \Cref{eq:l2} is the unweighted sum of three \(L_1\) terms: blood velocity, Doppler power, and the local velocity standard deviation (the velocity variation, computed in a \(3\times3\) neighborhood with neighbor differences wrapped into the Nyquist interval). The velocity supervision mask selects pixels inside the box that lie in dark B-mode regions (B-mode amplitude below 0.4 on a $[0,1]$ scale) and have Doppler power above 0.3, with a floor of 0.01 on the remaining box pixels to keep gradients flowing; pixels outside the box enter the velocity and power terms with weight 0.01 and a target of zero, which discourages very large velocities in areas where color boxes are seldom placed. The Segmentation loss is a foreground-only soft Dice loss with smoothing \(\varepsilon=10^{-6}\), computed over the whole clip for the LV cavity and the myocardium and averaged over the structures present. Training masks are rasterized from the strain contours on every frame, using the nearest annotated contour in time. When training on beamspace input, each row of the loss is weighted by a factor growing linearly along the radial axis (from \(0.001\) close to the probe to \(1.999\) at maximum depth), counteracting the depth-dependent pixel area in beamspace and approximately matching the pixel-area weighting that is implicit when training on Cartesian input.

The Task~4 model uses a cosine noise schedule~\cite{nichol_improved_2021} with 1000 steps and is trained to predict the velocity $v$ of the diffusion process~\cite{salimans_progressive_2022}, with the mean squared error on $v$ as loss. Generation uses deterministic DDIM sampling~\cite{song_denoising_2022} ($\eta=0$) with 32 steps.

\begin{table}[ht]
    \centering
    \caption{Per-task losses and masking.}
    \label{tab:losses}
    \footnotesize
    \setlength{\tabcolsep}{6pt}
    \begin{tabular*}{\linewidth}{@{\extracolsep{\fill}}lll@{}}
        \toprule
        \textbf{Task} & \textbf{Terms} & \textbf{Mask / weighting} \\
        \midrule
        Tissue Doppler & \Cref{eq:l1} & Pixels inside the sector \\
        Color Doppler  & \Cref{eq:l2} & Box \(\cap\) \{$B<0.4$\} \(\cap\) \{$P>0.3$\}, floor 0.01; outside-box weight 0.01 \\
        Segmentation   & \Cref{eq:l3} & Beamspace only: row weight \(0.001 \to 1.999\) along radial axis \\
        Video generation & MSE on \(v\) & None \\
        \bottomrule
    \end{tabular*}
    \par\vspace{2pt}
    \begin{minipage}{\linewidth}
    \footnotesize
    \(B\) and \(P\) are the B-mode amplitude and Doppler power.
    \end{minipage}
\end{table}

\subsection{Data filtering}
We keep only acquisitions whose timing and dynamic range are consistent with the supervised target, details are listed in \Cref{tab:data-filtering}. Tissue Doppler frames are assigned to the B-mode frame intervals by taking, for each interval, the Doppler frame closest to its midpoint, and clips with an empty interval are discarded. Color Doppler targets are placed at one quarter and three quarters of each B-mode frame interval, using two Doppler frames per interval when the B-mode to Doppler frame-count ratio is at most 0.75 and one repeated Doppler frame otherwise. No augmentations are applied in Tasks~1--3; Task~4 flips the ECG polarity with probability 0.5 during training, simulating differing placement of electrodes.

For Task~4, frames are scaled to $[0,1]$, bilinearly resized to the square resolution in beamspace, and mapped to $[-1,1]$. Scan-converted frames are produced from the resized beamspace frames by linear scan conversion onto a square grid spanning depths from zero to the maximum sector depth axially and $\pm$half that depth laterally, with zeros outside the sector.

\begin{table}[ht]
    \centering
    \caption{Per-task data filtering and clip sampling.
    Training clips are 32 frames long in every task.}
    \label{tab:data-filtering}
    \footnotesize
    \setlength{\tabcolsep}{6pt}
    \begin{tabular*}{\linewidth}{@{\extracolsep{\fill}}>{\raggedright\arraybackslash}p{2.2cm} >{\raggedright\arraybackslash}p{4.9cm} >{\raggedright\arraybackslash}p{5.3cm}@{}}
        \toprule
        \textbf{Task} & \textbf{Recording filters} & \textbf{Clip sampling} \\
        \midrule
        Tissue Doppler &
        B-mode FPS \(\in [26, 33]\) Hz; at least 33 B-mode frames; clips crossing a clock reset or scale change discarded &
        Training: stride 16, sliding. Evaluation: central 96 frames per recording. \\
        \addlinespace
        Color Doppler &
        Doppler:B-mode frame ratio \(\in [0.45, 1.1]\); B-mode FPS \(\in [9.0, 13.0]\) Hz; Nyquist~\(\in [0.60, 0.61]\) m/s &
        Training: stride 16, sliding. Evaluation: central window per recording. \\
        \addlinespace
        Segmentation &
        All left-ventricular strain recordings with contour annotations &
        Training: stride 32, sliding. Evaluation: annotated frames with 8 frames of context on each side. \\
        \addlinespace
        Video generation &
        A2C, A3C, A4C; B-mode FPS \(\ge 25\) Hz, every second frame above 50 Hz; clip within the ECG; sector starting at the probe &
        Training: one clip per recording, random start. \\
        \bottomrule
    \end{tabular*}
    \par\vspace{2pt}
    \begin{minipage}{\linewidth}
    \footnotesize
    \end{minipage}
\end{table}

\subsection{Cross-validation, evaluation, and reporting}
\paragraph{Task 1-3} Five-fold patient-level cross-validation is performed for every combination of method, input domain, input resolution, and task. Folds are fixed in advance and reused across all cells, so that beamspace and scan-converted variants of the same model see the same split. Validation metrics are always computed in the Cartesian domain, regardless of the input domain: when a model is trained on beamspace, its predictions are scan-converted to Cartesian (bilinear, zero outside the sector) before metric computation, and Task~3 probabilities are resampled before the argmax. This ensures that beamspace and scan-converted rows are scored against the same spatial reference. Evaluation uses sliding windows with $50\%$ overlap, averaged where they overlap.

The reported metrics are, for Task~1, the Nyquist-wrapped absolute velocity error over the sector in cm/s; for Task~2, the absolute errors of blood velocity (in cm/s, weighted by the velocity mask inside the box), Doppler power, and velocity variation (in cm/s); and for Task~3, the Dice score of the thresholded (argmax) segmentation over the annotated frames, averaged over the LV cavity and the myocardium. Within each fold, clip metrics are averaged per examination and then over examinations. \Cref{tab:results-r128,tab:results-r256,tab:results-r320} report the mean and sample standard deviation of these fold values across the five validation folds.

The domain comparison in \Cref{tab:domain-comparison-r128,tab:domain-comparison-r256,tab:domain-comparison-r320} pools the out-of-fold predictions of each method across the five folds, averages each metric per examination, and takes the per-examination difference between the beamspace and scan-converted variants. A cluster bootstrap over examinations (2000 replicates) gives the percentile 95\% confidence interval and a recentered two-sided $p$-value, and the Holm procedure~\cite{holm_simple_1979} corrects across the ten trained methods within each task metric and resolution.

\paragraph{Task 4} The model is trained on four folds and evaluated on the held-out fold, for each input domain and each of the three resolutions. One clip per recording is extracted, and the beamspace and scan-converted models are generated from the same clips with the same initial noise. Reference clips and beamspace outputs are scan-converted, and scan-converted outputs resized, onto the square Cartesian canvas of each feature extractor (with the physical extent used in training) and quantized to 8 bits, so that both arms are scored in one common image space.

Generation quality is measured in seven feature spaces (\Cref{tab:task4-extractors}). For the four frame-level extractors, the Fréchet distance (FD)~\cite{heusel_gans_2018} between generated and reference frame embeddings is computed per view with unbiased covariance estimates. For the three clip-level extractors, a paired unbiased kernel distance (KD)~\cite{binkowski_demystifying_2021} is computed per view with the cubic polynomial kernel $k(x,y)=(x^\top y/d+1)^3$.

\begin{table}[ht]
    \centering
    \caption{Feature extractors for the generation metrics.}
    \label{tab:task4-extractors}
    \footnotesize
    \setlength{\tabcolsep}{5pt}
    \begin{tabular*}{\linewidth}{@{\extracolsep{\fill}}llrrr@{}}
        \toprule
        \textbf{Feature space} & \textbf{Backbone} & \textbf{Frames} & \textbf{Input} & \textbf{Dimension} \\
        \midrule
        \multicolumn{5}{@{}l}{\textbf{Frame-level (FD)}} \\
        \quad EchoPrime view~\cite{vukadinovic_comprehensive_2026} & ConvNeXt-B & 1 & 224 px & 1024 \\
        \quad EchoFine (TTE47)~\cite{naidoo_robust_2026} & Xception & 1 & 224 px & 2048 \\
        \quad EchoCLIP~\cite{christensen_visionlanguage_2024} & ConvNeXt-B & 1 & 224 px & 512 \\
        \quad Inception (FID)~\cite{szegedy_rethinking_2016} & Inception-v3 & 1 & 299 px & 2048 \\
        \multicolumn{5}{@{}l}{\textbf{Clip-level (KD)}} \\
        \quad PanEcho~\cite{holste_complete_2025} & ConvNeXt-T + transformer & 16 (stride 1) & 224 px & 768 \\
        \quad EchoJEPA~\cite{munim_echojepa_2026} & ViT-L & 16 (stride 2) & 256 px & 1024 \\
        \quad EchoPrime MViT~\cite{vukadinovic_comprehensive_2026} & MViT-v2-S & 16 (stride 2) & 224 px & 512 \\
        \bottomrule
    \end{tabular*}
    \par\vspace{2pt}
    \begin{minipage}{\linewidth}
    \footnotesize
    Input is the network input size; clip-level extractors take a center crop of a larger canvas (PanEcho 256\,px, EchoJEPA 292\,px, EchoPrime 280\,px).
    \end{minipage}
\end{table}

Within each fold, the beamspace and scan-converted models are compared by a paired cluster bootstrap over examinations (2000 replicates) to obtain a two-sided $p$-value. \Cref{tab:task4-generation-cv} reports the mean of each metric over the five folds and the relative difference of the fold means. Its $p$-values come from a fold-stratified bootstrap, in which each of the 2000 pooled replicates averages one per-fold replicate from each fold and Holm-corrected~\cite{holm_simple_1979} across the 21 comparisons (seven feature spaces, three resolutions).

\subsection{Hardware and software}
\Cref{tab:hw-sw} lists the hardware and software stack used to produce the results of all four tasks. Each model was trained and evaluated on a single GPU. In total, all experiments presented took about 221 GPU hours, about 74 hours of wall-clock time with the three GPUs of \Cref{tab:hw-sw} fully occupied.

\begin{table}[ht]
    \centering
    \caption{Hardware and software environment used for training and evaluation.}
    \label{tab:hw-sw}
    \small
    \setlength{\tabcolsep}{4pt}
    \begin{tabular*}{\linewidth}{@{\extracolsep{\fill}}ll@{}}
        \toprule
        \textbf{Component} & \textbf{Version / model} \\
        \midrule
        GPU              & $3\times$\,NVIDIA GeForce RTX 5090 (32 GiB) \\
        CPU              & AMD Ryzen Threadripper PRO 9965WX (24 cores) \\
        System memory    & 188 GiB \\
        CUDA runtime     & 13.0 (PyTorch cu130 build) \\
        Python           & 3.13.7 \\
        PyTorch          & 2.12.0 \\
        timm             & 1.0.30 \\
        segmentation\_models\_pytorch & 0.5.0 \\
        \bottomrule
    \end{tabular*}
\end{table}

\subsection{Third-party assets and licenses}
DINOv3 ViT-S/16 denotes the architecture as implemented in timm. CAMUS~\cite{leclerc_deep_2019} is distributed under CC BY-NC-SA 4.0 and is used for evaluation only. The Task~4 feature extractors are used with their published weights: EchoCLIP (MIT), EchoPrime (MIT), PanEcho (weights CC BY-NC-SA 4.0, code AGPL-3.0), EchoJEPA (MIT with Apache-2.0 components, following V-JEPA~2), the torch-fidelity Inception-v3 FID weights (Apache 2.0), and the EchoFine TTE47 view classifier (MIT). Software: PyTorch (BSD-3-Clause), timm (Apache 2.0), segmentation\_models\_pytorch (MIT), torch-fidelity (Apache 2.0).

\clearpage
\section{Additional benchmark results}\label{app:additional-results}
\subsection{Other resolutions and per-method comparisons}
This part complements the $256\times256$ results of the main text. \Cref{tab:results-r128,tab:results-r320} report the Tasks~1--3 benchmark at $128\times128$ and $320\times320$, \Cref{tab:task4-generation-cv} reports Task~4 at all three resolutions, and \Cref{tab:domain-comparison-r128,tab:domain-comparison-r256,tab:domain-comparison-r320} break the Tasks~1--3 comparison down per method.

\begin{table}[ht]
    \centering
    \caption{Cross-validated benchmark results for Tasks 1-3 at $128\times128$ resolution.}
    \label{tab:results-r128}
    \footnotesize
    \setlength{\tabcolsep}{2.7pt}
    \sisetup{detect-weight=true, detect-family=true}
    \begin{tabular*}{\linewidth}{@{\extracolsep{\fill}}ll!{\extracolsep{\fill}}S[table-format=1.2]@{\extracolsep{0pt}\,$\pm$\,}S[table-format=1.2]!{\extracolsep{\fill}}S[table-format=2.2]@{\extracolsep{0pt}\,$\pm$\,}S[table-format=1.2]!{\extracolsep{\fill}}S[table-format=1.3]@{\extracolsep{0pt}\,$\pm$\,}S[table-format=1.3]!{\extracolsep{\fill}}S[table-format=1.2]@{\extracolsep{0pt}\,$\pm$\,}S[table-format=1.2]!{\extracolsep{\fill}}S[table-format=2.1]@{\extracolsep{0pt}\,$\pm$\,}S[table-format=1.1]@{}}
    \toprule
    \textbf{Model} & \textbf{Input domain} & \multicolumn{2}{c}{\textbf{Task 1} $\downarrow$} & \multicolumn{6}{c}{\textbf{Task 2} $\downarrow$} & \multicolumn{2}{c}{\textbf{Task 3} $\uparrow$} \\
    \cmidrule(lr){3-4}\cmidrule(lr){5-10}\cmidrule(lr){11-12}
    & & \multicolumn{2}{c}{\textbf{Velocity}} & \multicolumn{2}{c}{\textbf{Velocity}} & \multicolumn{2}{c}{\textbf{Power}} & \multicolumn{2}{c}{\textbf{Variation}} & \multicolumn{2}{c}{\textbf{Dice (\%)}} \\
    \midrule
    \multicolumn{12}{@{}l}{\textbf{Baseline}} \\
    \quad Temporal mean & Beamspace & 2.38 & 0.12 & 16.48 & 0.26 & 0.174 & 0.006 & 4.47 & 0.10 & 50.8 & 0.5 \\
     & Scan-converted & 2.38 & 0.12 & 16.48 & 0.26 & 0.175 & 0.006 & 4.58 & 0.10 & 51.3 & 0.6 \\
    \multicolumn{12}{@{}l}{\textbf{Framewise}} \\
    \quad 2D U-Net & Beamspace & 2.26 & 0.11 & 15.82 & 0.25 & 0.154 & 0.002 & 3.20 & 0.12 & 82.3 & 0.9 \\
     & Scan-converted & 2.31 & 0.13 & 15.78 & 0.23 & 0.153 & 0.002 & 3.19 & 0.11 & 82.9 & 0.8 \\
    \addlinespace[1pt]
    \quad ConvNeXt V2 Atto & Beamspace & 2.28 & 0.14 & 15.98 & 0.28 & 0.154 & 0.004 & 3.29 & 0.13 & 82.6 & 0.9 \\
     & Scan-converted & 2.29 & 0.10 & 16.00 & 0.29 & 0.153 & 0.003 & 3.26 & 0.15 & 82.7 & 1.0 \\
    \addlinespace[1pt]
    \quad SegFormer-B0 & Beamspace & 2.32 & 0.16 & 15.89 & 0.27 & 0.156 & 0.003 & 3.42 & 0.20 & 83.0 & 1.0 \\
     & Scan-converted & 2.28 & 0.10 & 15.89 & 0.29 & 0.155 & 0.005 & 3.42 & 0.34 & 83.2 & 0.8 \\
    \addlinespace[1pt]
    \quad DINOv3 ViT-S/16 & Beamspace & 2.25 & 0.11 & 15.82 & 0.19 & 0.152 & 0.003 & 3.19 & 0.10 & 82.0 & 0.9 \\
     & Scan-converted & 2.31 & 0.11 & 16.16 & 0.32 & 0.154 & 0.004 & 3.21 & 0.12 & 81.7 & 0.7 \\
    \addlinespace[1pt]
    \quad MambaOut-Femto & Beamspace & 2.23 & 0.12 & 15.69 & 0.13 & 0.152 & 0.003 & 3.19 & 0.13 & 83.1 & 0.8 \\
     & Scan-converted & 2.25 & 0.11 & 15.92 & 0.26 & 0.155 & 0.003 & 3.18 & 0.13 & 83.3 & 0.9 \\
    \multicolumn{12}{@{}l}{\textbf{Time-aware}} \\
    \quad 3D U-Net & Beamspace & 1.29 & 0.07 & 14.25 & 0.37 & 0.129 & 0.003 & 3.14 & 0.11 & 82.6 & 0.9 \\
     & Scan-converted & 1.31 & 0.07 & 14.36 & 0.40 & 0.134 & 0.003 & 3.16 & 0.15 & 83.3 & 1.0 \\
    \addlinespace[1pt]
    \quad Video Swin-T & Beamspace & 1.41 & 0.08 & 14.56 & 0.30 & 0.131 & 0.002 & 3.18 & 0.09 & 82.7 & 1.0 \\
     & Scan-converted & 1.42 & 0.06 & 14.57 & 0.36 & 0.133 & 0.003 & 3.14 & 0.12 & 83.2 & 0.8 \\
    \addlinespace[1pt]
    \quad MViT-v2-S & Beamspace & 1.31 & 0.07 & \bfseries 13.98 & 0.32 & 0.125 & 0.003 & 3.14 & 0.06 & 83.5 & 0.7 \\
     & Scan-converted & 1.32 & 0.08 & 14.13 & 0.28 & 0.126 & 0.003 & 3.14 & 0.11 & 83.6 & 1.0 \\
    \addlinespace[1pt]
    \quad UniFormer-S 32f & Beamspace & 1.32 & 0.07 & 14.17 & 0.36 & 0.127 & 0.003 & 3.15 & 0.11 & 83.1 & 0.8 \\
     & Scan-converted & 1.34 & 0.08 & 14.17 & 0.36 & 0.129 & 0.002 & \bfseries 3.13 & 0.12 & \bfseries 83.7 & 0.8 \\
    \addlinespace[1pt]
    \quad U-Mamba-Bot & Beamspace & \bfseries 1.26 & 0.07 & 14.10 & 0.32 & \bfseries 0.124 & 0.002 & 3.17 & 0.07 & 83.2 & 0.6 \\
     & Scan-converted & 1.28 & 0.07 & 14.07 & 0.40 & 0.125 & 0.002 & 3.15 & 0.07 & 83.6 & 0.9 \\
    \bottomrule
    \end{tabular*}
    \par\vspace{1pt}
    \begin{minipage}{\linewidth}
    \footnotesize
    Mean $\pm$ standard deviation across the folds. $\downarrow$ lower is better, $\uparrow$ higher is better. Velocities and variation in \SI{}{\centi\meter/\second}.
    \end{minipage}
\end{table}

\begin{table}[ht]
    \centering
    \caption{Cross-validated benchmark results for Tasks 1-3 at $320\times320$ resolution.}
    \label{tab:results-r320}
    \footnotesize
    \setlength{\tabcolsep}{2.7pt}
    \sisetup{detect-weight=true, detect-family=true}
    \begin{tabular*}{\linewidth}{@{\extracolsep{\fill}}ll!{\extracolsep{\fill}}S[table-format=1.2]@{\extracolsep{0pt}\,$\pm$\,}S[table-format=1.2]!{\extracolsep{\fill}}S[table-format=2.2]@{\extracolsep{0pt}\,$\pm$\,}S[table-format=1.2]!{\extracolsep{\fill}}S[table-format=1.3]@{\extracolsep{0pt}\,$\pm$\,}S[table-format=1.3]!{\extracolsep{\fill}}S[table-format=1.2]@{\extracolsep{0pt}\,$\pm$\,}S[table-format=1.2]!{\extracolsep{\fill}}S[table-format=2.1]@{\extracolsep{0pt}\,$\pm$\,}S[table-format=1.1]@{}}
    \toprule
    \textbf{Model} & \textbf{Input domain} & \multicolumn{2}{c}{\textbf{Task 1} $\downarrow$} & \multicolumn{6}{c}{\textbf{Task 2} $\downarrow$} & \multicolumn{2}{c}{\textbf{Task 3} $\uparrow$} \\
    \cmidrule(lr){3-4}\cmidrule(lr){5-10}\cmidrule(lr){11-12}
    & & \multicolumn{2}{c}{\textbf{Velocity}} & \multicolumn{2}{c}{\textbf{Velocity}} & \multicolumn{2}{c}{\textbf{Power}} & \multicolumn{2}{c}{\textbf{Variation}} & \multicolumn{2}{c}{\textbf{Dice (\%)}} \\
    \midrule
    \multicolumn{12}{@{}l}{\textbf{Baseline}} \\
    \quad Temporal mean & Beamspace & 2.38 & 0.12 & 16.47 & 0.26 & 0.174 & 0.006 & 3.32 & 0.07 & 50.7 & 0.6 \\
     & Scan-converted & 2.38 & 0.12 & 16.47 & 0.26 & 0.176 & 0.006 & 3.57 & 0.07 & 51.1 & 0.7 \\
    \multicolumn{12}{@{}l}{\textbf{Framewise}} \\
    \quad 2D U-Net & Beamspace & 2.19 & 0.10 & 15.66 & 0.25 & 0.154 & 0.002 & 2.14 & 0.05 & 82.9 & 0.8 \\
     & Scan-converted & 2.17 & 0.11 & 15.72 & 0.20 & 0.152 & 0.003 & 2.13 & 0.06 & 83.3 & 0.8 \\
    \addlinespace[1pt]
    \quad ConvNeXt V2 Atto & Beamspace & 2.18 & 0.12 & 15.71 & 0.23 & 0.152 & 0.003 & 2.20 & 0.06 & 83.2 & 1.0 \\
     & Scan-converted & 2.31 & 0.14 & 15.84 & 0.19 & 0.155 & 0.003 & 2.14 & 0.08 & 83.6 & 0.6 \\
    \addlinespace[1pt]
    \quad SegFormer-B0 & Beamspace & 2.20 & 0.09 & 15.53 & 0.25 & 0.152 & 0.001 & 2.23 & 0.12 & 83.5 & 0.7 \\
     & Scan-converted & 2.33 & 0.17 & 15.94 & 0.47 & 0.156 & 0.003 & 2.28 & 0.28 & 84.0 & 0.8 \\
    \addlinespace[1pt]
    \quad DINOv3 ViT-S/16 & Beamspace & 2.18 & 0.12 & 15.53 & 0.21 & 0.151 & 0.003 & 2.19 & 0.12 & 83.3 & 0.7 \\
     & Scan-converted & 2.19 & 0.09 & 15.62 & 0.21 & 0.152 & 0.003 & 2.13 & 0.08 & 84.0 & 0.9 \\
    \addlinespace[1pt]
    \quad MambaOut-Femto & Beamspace & 2.26 & 0.21 & 15.48 & 0.21 & 0.152 & 0.003 & 2.17 & 0.07 & 83.3 & 0.7 \\
     & Scan-converted & 2.30 & 0.18 & 15.61 & 0.31 & 0.153 & 0.004 & 2.13 & 0.05 & 83.9 & 0.7 \\
    \multicolumn{12}{@{}l}{\textbf{Time-aware}} \\
    \quad 3D U-Net & Beamspace & 1.23 & 0.07 & 14.10 & 0.29 & 0.127 & 0.003 & 2.10 & 0.06 & 83.2 & 0.8 \\
     & Scan-converted & 1.23 & 0.06 & 14.10 & 0.42 & 0.126 & 0.002 & 2.11 & 0.08 & 83.7 & 0.8 \\
    \addlinespace[1pt]
    \quad Video Swin-T & Beamspace & 1.38 & 0.09 & 14.51 & 0.27 & 0.132 & 0.002 & 2.18 & 0.03 & 83.0 & 0.7 \\
     & Scan-converted & 1.40 & 0.08 & 14.59 & 0.30 & 0.133 & 0.003 & 2.10 & 0.07 & 83.6 & 0.9 \\
    \addlinespace[1pt]
    \quad MViT-v2-S & Beamspace & 1.25 & 0.07 & \bfseries 13.82 & 0.31 & 0.124 & 0.002 & 2.10 & 0.07 & 83.7 & 0.9 \\
     & Scan-converted & 1.27 & 0.07 & 13.94 & 0.35 & 0.125 & 0.002 & 2.10 & 0.06 & \bfseries 84.6 & 0.8 \\
    \addlinespace[1pt]
    \quad UniFormer-S 32f & Beamspace & 1.27 & 0.07 & 13.99 & 0.31 & 0.127 & 0.002 & 2.12 & 0.08 & 83.8 & 0.7 \\
     & Scan-converted & 1.29 & 0.07 & 14.01 & 0.29 & 0.129 & 0.002 & 2.13 & 0.08 & 84.5 & 0.6 \\
    \addlinespace[1pt]
    \quad U-Mamba-Bot & Beamspace & \bfseries 1.18 & 0.07 & 14.04 & 0.38 & \bfseries 0.122 & 0.003 & 2.11 & 0.08 & 83.2 & 0.7 \\
     & Scan-converted & 1.20 & 0.06 & 13.95 & 0.35 & 0.123 & 0.003 & \bfseries 2.09 & 0.06 & 84.0 & 0.6 \\
    \bottomrule
    \end{tabular*}
    \par\vspace{1pt}
    \begin{minipage}{\linewidth}
    \footnotesize
    Mean $\pm$ standard deviation across the folds. $\downarrow$ lower is better, $\uparrow$ higher is better. Velocities and variation in \SI{}{\centi\meter/\second}.
    \end{minipage}
\end{table}

\begin{table}[ht]
    \centering
    \caption{ECG-conditioned generation quality (Task~4) across resolutions, beamspace versus scan-converted training domain.}
    \label{tab:task4-generation-cv}
    \footnotesize
    \setlength{\tabcolsep}{3pt}
    \begin{tabular*}{\linewidth}{@{\extracolsep{\fill}}lcccc@{}}
    \toprule
    \textbf{Feature space} & \textbf{Beamspace} & \textbf{Scan-converted} & \textbf{Ratio} & \textbf{$p_\mathrm{Holm}$} \\
    \midrule
    \multicolumn{5}{@{}l}{\textbf{\boldmath$128\times128$}} \\
    \multicolumn{5}{@{}l}{\quad Frame-level (FD $\downarrow$)} \\
    \qquad EchoCLIP & $\mathbf{0.279} \pm 0.007$ & $0.313 \pm 0.005$ & $1.12\times$ & 0.01 \\
    \qquad EchoFine & $\mathbf{6.74} \pm 1.32$ & $12.1 \pm 1.6$ & $1.79\times$ & 0.01 \\
    \qquad EchoPrime view & $\mathbf{0.385} \pm 0.072$ & $0.600 \pm 0.120$ & $1.56\times$ & 0.01 \\
    \qquad Inception & $\mathbf{38.8} \pm 1.8$ & $138 \pm 4$ & $3.56\times$ & 0.01 \\
    \multicolumn{5}{@{}l}{\quad Clip-level (KD $\downarrow$)} \\
    \qquad EchoJEPA & $\mathbf{36.4} \pm 0.8$ & $66.5 \pm 1.6$ & $1.83\times$ & 0.01 \\
    \qquad EchoPrime MViT & $7.06 \pm 0.46$ & $\mathbf{5.57} \pm 0.19$ & $0.79\times$ & 0.01 \\
    \qquad PanEcho & $\mathbf{0.382} \pm 0.018$ & $0.546 \pm 0.049$ & $1.43\times$ & 0.01 \\
    \addlinespace
    \multicolumn{5}{@{}l}{\textbf{\boldmath$256\times256$}} \\
    \multicolumn{5}{@{}l}{\quad Frame-level (FD $\downarrow$)} \\
    \qquad EchoCLIP & $\mathbf{0.163} \pm 0.008$ & $0.235 \pm 0.005$ & $1.44\times$ & 0.01 \\
    \qquad EchoFine & $\mathbf{5.35} \pm 0.89$ & $7.54 \pm 1.77$ & $1.41\times$ & 0.01 \\
    \qquad EchoPrime view & $\mathbf{0.373} \pm 0.090$ & $0.451 \pm 0.103$ & $1.21\times$ & 0.016 \\
    \qquad Inception & $\mathbf{14.8} \pm 0.3$ & $39.8 \pm 1.3$ & $2.69\times$ & 0.01 \\
    \multicolumn{5}{@{}l}{\quad Clip-level (KD $\downarrow$)} \\
    \qquad EchoJEPA & $\mathbf{6.54} \pm 0.99$ & $27.0 \pm 0.6$ & $4.12\times$ & 0.01 \\
    \qquad EchoPrime MViT & $\mathbf{1.76} \pm 0.13$ & $5.60 \pm 0.16$ & $3.19\times$ & 0.01 \\
    \qquad PanEcho & $\mathbf{0.209} \pm 0.018$ & $0.234 \pm 0.015$ & $1.12\times$ & 0.01 \\
    \addlinespace
    \multicolumn{5}{@{}l}{\textbf{\boldmath$320\times320$}} \\
    \multicolumn{5}{@{}l}{\quad Frame-level (FD $\downarrow$)} \\
    \qquad EchoCLIP & $\mathbf{0.134} \pm 0.004$ & $0.191 \pm 0.006$ & $1.43\times$ & 0.01 \\
    \qquad EchoFine & $\mathbf{4.99} \pm 1.16$ & $6.46 \pm 1.91$ & $1.29\times$ & 0.01 \\
    \qquad EchoPrime view & $\mathbf{0.399} \pm 0.068$ & $0.465 \pm 0.148$ & $1.16\times$ & 0.025 \\
    \qquad Inception & $\mathbf{12.9} \pm 0.5$ & $27.4 \pm 1.4$ & $2.13\times$ & 0.01 \\
    \multicolumn{5}{@{}l}{\quad Clip-level (KD $\downarrow$)} \\
    \qquad EchoJEPA & $\mathbf{5.20} \pm 0.53$ & $13.5 \pm 1.0$ & $2.59\times$ & 0.01 \\
    \qquad EchoPrime MViT & $\mathbf{1.20} \pm 0.04$ & $3.59 \pm 0.28$ & $2.99\times$ & 0.01 \\
    \qquad PanEcho & $\mathbf{0.198} \pm 0.026$ & $0.352 \pm 0.038$ & $1.78\times$ & 0.01 \\
    \bottomrule
    \end{tabular*}
    \par\vspace{2pt}
    \begin{minipage}{\linewidth}
    \footnotesize
    Mean $\pm$ standard deviation across the five validation folds, each fold averaged over views. Ratio: scan-converted / beamspace mean (above 1 favors beamspace). Bold: the better domain where $p_\mathrm{Holm}<0.05$.\\
    $p$: fold-stratified paired examination-level bootstrap (2000 replicates per fold), recentered, two-sided, Holm-adjusted across all 21 comparisons; the smallest attainable value is 0.0105, shown as 0.01.
    \end{minipage}
\end{table}

\begin{table}[ht]
    \centering
    \caption{Beamspace versus scan-converted input at $128\times128$, per method and task metric.}
    \label{tab:domain-comparison-r128}
    \footnotesize
    \setlength{\tabcolsep}{4pt}
    \sisetup{detect-weight=true, detect-family=true}
    \begin{tabular*}{\linewidth}{@{\extracolsep{\fill}}l*{5}{!{\extracolsep{\fill}}r@{\extracolsep{0pt}.}l}@{}}
    \toprule
    & \multicolumn{2}{c}{\textbf{Task 1}} & \multicolumn{6}{c}{\textbf{Task 2}} & \multicolumn{2}{c}{\textbf{Task 3}} \\
    \cmidrule(lr){2-3} \cmidrule(lr){4-9} \cmidrule(lr){10-11}
    \textbf{Method} & \multicolumn{2}{c}{\textbf{Velocity}} & \multicolumn{2}{c}{\textbf{Velocity}} & \multicolumn{2}{c}{\textbf{Power}} & \multicolumn{2}{c}{\textbf{Variation}} & \multicolumn{2}{c}{\textbf{Dice}} \\
    \midrule
    \multicolumn{11}{@{}l}{\textbf{Framewise}} \\
    \quad 2D U-Net & +2 & 5\%$^{**}$ & $-$0 & 2\% & $-$0 & 4\% & $-$0 & 2\% & $-$0 & 8\%$^{**}$ \\
    \quad ConvNeXt V2 Atto & +0 & 3\% & +0 & 1\% & $-$0 & 4\% & $-$0 & 9\% & $-$0 & 1\% \\
    \quad SegFormer-B0 & $-$1 & 8\%$^{**}$ & $-$0 & 1\% & $-$0 & 6\% & +0 & 1\% & $-$0 & 4\% \\
    \quad DINOv3 ViT-S/16 & +2 & 6\%$^{**}$ & +2 & 0\%$^{**}$ & +1 & 5\%$^{**}$ & +0 & 5\% & +0 & 4\% \\
    \quad MambaOut-Femto & +1 & 2\%$^{*}$ & +1 & 4\%$^{**}$ & +1 & 7\%$^{*}$ & $-$0 & 3\% & $-$0 & 2\% \\
    \addlinespace
    \multicolumn{11}{@{}l}{\textbf{Time-aware}} \\
    \quad 3D U-Net & +1 & 3\%$^{**}$ & +0 & 8\%$^{*}$ & +3 & 5\%$^{**}$ & +0 & 6\% & $-$0 & 7\%$^{*}$ \\
    \quad Video Swin-T & +1 & 1\%$^{**}$ & 0 & 0\% & +1 & 7\%$^{**}$ & $-$1 & 2\%$^{**}$ & $-$0 & 6\%$^{*}$ \\
    \quad MViT-v2-S & +1 & 0\%$^{**}$ & +1 & 0\%$^{**}$ & +1 & 4\%$^{**}$ & +0 & 1\% & $-$0 & 1\% \\
    \quad UniFormer-S 32f & +1 & 9\%$^{**}$ & +0 & 1\% & +1 & 4\%$^{**}$ & $-$0 & 6\%$^{*}$ & $-$0 & 7\%$^{*}$ \\
    \quad U-Mamba-Bot & +1 & 4\%$^{**}$ & $-$0 & 2\% & +1 & 0\% & $-$0 & 8\%$^{*}$ & $-$0 & 5\% \\
    \bottomrule
    \end{tabular*}
    \par\vspace{2pt}
    \begin{minipage}{\linewidth}
    \footnotesize
Relative difference between the beamspace and scan-converted variants, signed so that a positive value favours beamspace (lower error in Tasks~1 and~2, higher Dice in Task~3). $^{*}p_\mathrm{Holm}<0.05$, $^{**}p_\mathrm{Holm}<0.01$; paired examination-level bootstrap, two-sided, Holm-adjusted within each task metric.
    \end{minipage}
\end{table}

\begin{table}[ht]
    \centering
    \caption{Beamspace versus scan-converted input at $256\times256$, per method and task metric.}
    \label{tab:domain-comparison-r256}
    \footnotesize
    \setlength{\tabcolsep}{4pt}
    \sisetup{detect-weight=true, detect-family=true}
    \begin{tabular*}{\linewidth}{@{\extracolsep{\fill}}l*{5}{!{\extracolsep{\fill}}r@{\extracolsep{0pt}.}l}@{}}
    \toprule
    & \multicolumn{2}{c}{\textbf{Task 1}} & \multicolumn{6}{c}{\textbf{Task 2}} & \multicolumn{2}{c}{\textbf{Task 3}} \\
    \cmidrule(lr){2-3} \cmidrule(lr){4-9} \cmidrule(lr){10-11}
    \textbf{Method} & \multicolumn{2}{c}{\textbf{Velocity}} & \multicolumn{2}{c}{\textbf{Velocity}} & \multicolumn{2}{c}{\textbf{Power}} & \multicolumn{2}{c}{\textbf{Variation}} & \multicolumn{2}{c}{\textbf{Dice}} \\
    \midrule
    \multicolumn{11}{@{}l}{\textbf{Framewise}} \\
    \quad 2D U-Net & $-$0 & 2\% & +0 & 4\% & +0 & 2\% & $-$0 & 9\%$^{**}$ & $-$0 & 6\%$^{*}$ \\
    \quad ConvNeXt V2 Atto & +3 & 1\%$^{**}$ & +0 & 9\%$^{**}$ & +0 & 7\% & $-$2 & 3\%$^{**}$ & $-$0 & 5\%$^{*}$ \\
    \quad SegFormer-B0 & +5 & 6\%$^{**}$ & $-$1 & 1\%$^{**}$ & +1 & 5\% & $-$12 & 8\%$^{**}$ & $-$0 & 4\%$^{*}$ \\
    \quad DINOv3 ViT-S/16 & +0 & 9\% & +0 & 7\%$^{**}$ & +1 & 6\%$^{*}$ & $-$0 & 8\%$^{*}$ & $-$0 & 5\%$^{*}$ \\
    \quad MambaOut-Femto & +1 & 5\%$^{**}$ & $-$0 & 1\% & +0 & 4\% & $-$1 & 9\%$^{**}$ & $-$0 & 9\%$^{**}$ \\
    \addlinespace
    \multicolumn{11}{@{}l}{\textbf{Time-aware}} \\
    \quad 3D U-Net & +0 & 2\% & $-$0 & 3\% & +0 & 5\% & +0 & 6\% & $-$0 & 6\%$^{*}$ \\
    \quad Video Swin-T & +1 & 5\%$^{**}$ & +0 & 4\% & +0 & 4\% & +0 & 4\% & $-$0 & 6\%$^{*}$ \\
    \quad MViT-v2-S & +1 & 2\%$^{**}$ & +0 & 8\%$^{*}$ & +0 & 9\% & 0 & 0\% & $-$0 & 8\%$^{**}$ \\
    \quad UniFormer-S 32f & +1 & 5\%$^{**}$ & +0 & 1\% & +2 & 0\%$^{**}$ & $-$0 & 5\% & $-$0 & 7\%$^{*}$ \\
    \quad U-Mamba-Bot & +1 & 7\%$^{**}$ & $-$0 & 2\% & +1 & 5\%$^{*}$ & 0 & 0\% & $-$0 & 6\%$^{*}$ \\
    \bottomrule
    \end{tabular*}
    \par\vspace{2pt}
    \begin{minipage}{\linewidth}
    \footnotesize
Relative difference between the beamspace and scan-converted variants, signed so that a positive value favours beamspace (lower error in Tasks~1 and~2, higher Dice in Task~3). $^{*}p_\mathrm{Holm}<0.05$, $^{**}p_\mathrm{Holm}<0.01$; paired examination-level bootstrap, two-sided, Holm-adjusted within each task metric.
    \end{minipage}
\end{table}

\begin{table}[ht]
    \centering
    \caption{Beamspace versus scan-converted input at $320\times320$, per method and task metric.}
    \label{tab:domain-comparison-r320}
    \footnotesize
    \setlength{\tabcolsep}{4pt}
    \sisetup{detect-weight=true, detect-family=true}
    \begin{tabular*}{\linewidth}{@{\extracolsep{\fill}}l*{5}{!{\extracolsep{\fill}}r@{\extracolsep{0pt}.}l}@{}}
    \toprule
    & \multicolumn{2}{c}{\textbf{Task 1}} & \multicolumn{6}{c}{\textbf{Task 2}} & \multicolumn{2}{c}{\textbf{Task 3}} \\
    \cmidrule(lr){2-3} \cmidrule(lr){4-9} \cmidrule(lr){10-11}
    \textbf{Method} & \multicolumn{2}{c}{\textbf{Velocity}} & \multicolumn{2}{c}{\textbf{Velocity}} & \multicolumn{2}{c}{\textbf{Power}} & \multicolumn{2}{c}{\textbf{Variation}} & \multicolumn{2}{c}{\textbf{Dice}} \\
    \midrule
    \multicolumn{11}{@{}l}{\textbf{Framewise}} \\
    \quad 2D U-Net & $-$0 & 8\% & +0 & 4\% & $-$1 & 0\% & $-$0 & 8\%$^{**}$ & $-$0 & 5\%$^{*}$ \\
    \quad ConvNeXt V2 Atto & +6 & 0\%$^{**}$ & +0 & 8\%$^{**}$ & +1 & 5\%$^{**}$ & $-$2 & 6\%$^{**}$ & $-$0 & 5\%$^{*}$ \\
    \quad SegFormer-B0 & +5 & 7\%$^{**}$ & +2 & 6\%$^{**}$ & +2 & 3\%$^{**}$ & +2 & 5\%$^{*}$ & $-$0 & 5\%$^{*}$ \\
    \quad DINOv3 ViT-S/16 & +0 & 4\% & +0 & 6\%$^{**}$ & +0 & 5\% & $-$2 & 6\%$^{**}$ & $-$0 & 8\%$^{**}$ \\
    \quad MambaOut-Femto & +1 & 7\%$^{**}$ & +0 & 7\% & +0 & 5\% & $-$2 & 4\%$^{**}$ & $-$0 & 8\%$^{**}$ \\
    \addlinespace
    \multicolumn{11}{@{}l}{\textbf{Time-aware}} \\
    \quad 3D U-Net & +0 & 2\% & $-$0 & 1\% & $-$0 & 1\% & +0 & 6\% & $-$0 & 6\%$^{*}$ \\
    \quad Video Swin-T & +1 & 5\%$^{**}$ & +0 & 6\% & +0 & 3\% & $-$3 & 5\%$^{**}$ & $-$0 & 8\%$^{**}$ \\
    \quad MViT-v2-S & +1 & 7\%$^{**}$ & +0 & 9\%$^{**}$ & +0 & 9\% & $-$0 & 3\% & $-$1 & 0\%$^{**}$ \\
    \quad UniFormer-S 32f & +1 & 5\%$^{**}$ & +0 & 1\% & +1 & 5\%$^{**}$ & +0 & 3\% & $-$0 & 9\%$^{**}$ \\
    \quad U-Mamba-Bot & +1 & 5\%$^{**}$ & $-$0 & 6\%$^{*}$ & +1 & 0\% & $-$1 & 0\%$^{**}$ & $-$1 & 0\%$^{**}$ \\
    \bottomrule
    \end{tabular*}
    \par\vspace{2pt}
    \begin{minipage}{\linewidth}
    \footnotesize
Relative difference between the beamspace and scan-converted variants, signed so that a positive value favors beamspace (lower error in Tasks~1 and~2, higher Dice in Task~3). $^{*}p_\mathrm{Holm}<0.05$, $^{**}p_\mathrm{Holm}<0.01$; paired examination-level bootstrap, two-sided, Holm-adjusted within each task metric.
    \end{minipage}
\end{table}

\clearpage
\subsection{External validation on CAMUS}
\Cref{tab:camus-segmentation,tab:task4-generation-camus} report the external validation of the Task~3 and Task~4 models on CAMUS~\cite{leclerc_deep_2019}, trained on EchoXFlow only. CAMUS is distributed scan-converted and records no scan geometry. Its beamspace input was obtained by resampling each frame onto a polar grid whose sector was fitted to the imaged fan, with cubic-spline interpolation at most half a source pixel apart; scan-converted back to the original grid. This reconstructed beamspace is an estimate of the acquisition geometry and differs from the native beamspace arrays of EchoXFlow. Without any adaptation, the segmentation models reach 73--80\% Dice on CAMUS against 82--85\% on the EchoXFlow folds, and the small but consistent advantage of scan-converted input seen in-distribution disappears: neither domain wins at every resolution. For Task~4, the beamspace models keep the better score in 14 of the 21 comparisons, including every clip-level metric at $256\times256$ and $320\times320$, the scan-converted models win 4, mostly frame-level metrics at $320\times320$, and 3 are not significant, so the in-distribution advantage of beamspace training largely transfers to CAMUS. The CAMUS clips are conditioned on the ECG of an EchoXFlow recording of the same view, so this experiment measures the transfer of generated image and video quality; the correspondence between a patient's own ECG and their cardiac motion is assessed in-distribution only.

\begin{table}[ht]
    \centering
    \caption{External validation of the Task~3 models on CAMUS: left-ventricular Dice (\%) on the end-diastolic and end-systolic frames of all 500 patients, A2C and A4C views, without any training on CAMUS.}
    \label{tab:camus-segmentation}
    \footnotesize
    \setlength{\tabcolsep}{2.7pt}
    \sisetup{detect-weight=true, detect-family=true}
    \begin{tabular*}{\linewidth}{@{\extracolsep{\fill}}ll!{\extracolsep{\fill}}S[table-format=2.1]@{\extracolsep{0pt}\,$\pm$\,}S[table-format=1.1]!{\extracolsep{\fill}}S[table-format=2.1]@{\extracolsep{0pt}\,$\pm$\,}S[table-format=1.1]!{\extracolsep{\fill}}S[table-format=2.1]@{\extracolsep{0pt}\,$\pm$\,}S[table-format=1.1]@{}}
    \toprule
    \textbf{Model} & \textbf{Input domain} & \multicolumn{2}{c}{\boldmath$128\times128$} & \multicolumn{2}{c}{\boldmath$256\times256$} & \multicolumn{2}{c}{\boldmath$320\times320$} \\
    \cmidrule(lr){3-4}\cmidrule(lr){5-6}\cmidrule(lr){7-8}
    & & \multicolumn{2}{c}{\textbf{Dice (\%)} $\uparrow$} & \multicolumn{2}{c}{\textbf{Dice (\%)} $\uparrow$} & \multicolumn{2}{c}{\textbf{Dice (\%)} $\uparrow$} \\
    \midrule
    \multicolumn{8}{@{}l}{\textbf{Baseline}} \\
    \quad Temporal mean & Beamspace & 45.5 & 0.3 & 45.5 & 0.2 & 45.4 & 0.3 \\
     & Scan-converted & 46.0 & 0.2 & 45.8 & 0.2 & 45.8 & 0.2 \\
    \multicolumn{8}{@{}l}{\textbf{Framewise}} \\
    \quad 2D U-Net & Beamspace & 77.3 & 0.9 & 77.5 & 0.5 & 76.9 & 0.5 \\
     & Scan-converted & 76.6 & 0.8 & 76.8 & 0.8 & 77.0 & 0.8 \\
    \addlinespace[1pt]
    \quad ConvNeXt V2 Atto & Beamspace & 77.2 & 0.7 & 78.3 & 0.5 & 78.0 & 0.5 \\
     & Scan-converted & 77.1 & 0.9 & 77.6 & 0.6 & 77.1 & 1.0 \\
    \addlinespace[1pt]
    \quad SegFormer-B0 & Beamspace & 78.6 & 1.0 & \bfseries 79.6 & 1.3 & 78.6 & 1.8 \\
     & Scan-converted & 78.5 & 0.6 & 78.5 & 0.6 & 78.6 & 0.3 \\
    \addlinespace[1pt]
    \quad DINOv3 ViT-S/16 & Beamspace & 76.9 & 0.7 & 78.0 & 0.3 & 77.9 & 0.7 \\
     & Scan-converted & 76.4 & 0.4 & 77.5 & 0.8 & 78.0 & 0.8 \\
    \addlinespace[1pt]
    \quad MambaOut-Femto & Beamspace & 78.4 & 0.3 & 78.0 & 0.5 & 78.4 & 0.5 \\
     & Scan-converted & 78.2 & 1.2 & 77.9 & 1.1 & 78.0 & 1.2 \\
    \multicolumn{8}{@{}l}{\textbf{Time-aware}} \\
    \quad 3D U-Net & Beamspace & 77.5 & 1.2 & 77.9 & 1.1 & 77.4 & 0.7 \\
     & Scan-converted & 76.8 & 0.7 & 76.8 & 0.9 & 77.0 & 0.6 \\
    \addlinespace[1pt]
    \quad Video Swin-T & Beamspace & 77.1 & 1.5 & 75.7 & 1.7 & 72.9 & 1.3 \\
     & Scan-converted & 76.7 & 1.1 & 76.6 & 1.2 & 76.9 & 0.6 \\
    \addlinespace[1pt]
    \quad MViT-v2-S & Beamspace & 79.2 & 0.8 & 78.8 & 0.3 & 79.4 & 0.8 \\
     & Scan-converted & \bfseries 79.3 & 0.9 & 79.1 & 0.9 & \bfseries 79.8 & 0.7 \\
    \addlinespace[1pt]
    \quad UniFormer-S 32f & Beamspace & 78.8 & 1.3 & 78.7 & 1.6 & 78.8 & 0.8 \\
     & Scan-converted & 78.4 & 0.9 & 78.3 & 0.8 & 79.2 & 1.0 \\
    \addlinespace[1pt]
    \quad U-Mamba-Bot & Beamspace & 78.5 & 1.0 & 77.5 & 0.9 & 78.1 & 0.7 \\
     & Scan-converted & 77.7 & 0.7 & 78.1 & 0.9 & 78.4 & 0.3 \\
    \bottomrule
    \end{tabular*}
    \par\vspace{1pt}
    \begin{minipage}{\linewidth}
    \footnotesize
    Mean $\pm$ standard deviation across the five cross-validation models of each cell, each model scored on the same 1000 CAMUS recordings (500 patients $\times$ 2 views) after the Task~3 preprocessing of its input domain (beamspace resize, or scan conversion then resize). Dice of the cavity and myocardium averaged over the ED and ES frames, the only CAMUS frames with labels.
    \end{minipage}
\end{table}

\begin{table}[ht]
    \centering
    \caption{Generation quality of the Task~4 models conditioned on CAMUS end-diastolic frames, beamspace versus scan-converted training domain.}
    \label{tab:task4-generation-camus}
    \footnotesize
    \setlength{\tabcolsep}{3pt}
    \begin{tabular*}{\linewidth}{@{\extracolsep{\fill}}lcccc@{}}
    \toprule
    \textbf{Feature space} & \textbf{Beamspace} & \textbf{Scan-converted} & \textbf{Ratio} & \textbf{$p_\mathrm{Holm}$} \\
    \midrule
    \multicolumn{5}{@{}l}{\textbf{\boldmath$128\times128$}} \\
    \multicolumn{5}{@{}l}{\quad Frame-level (FD $\downarrow$)} \\
    \qquad EchoCLIP & $\mathbf{0.251}$ & 0.350 & $1.40\times$ & 0.01 \\
    \qquad EchoFine & $\mathbf{5.50}$ & 10.1 & $1.84\times$ & 0.01 \\
    \qquad EchoPrime view & 0.626 & 0.579 & $0.92\times$ & 0.33 \\
    \qquad Inception & $\mathbf{50.2}$ & 108 & $2.15\times$ & 0.01 \\
    \multicolumn{5}{@{}l}{\quad Clip-level (KD $\downarrow$)} \\
    \qquad EchoJEPA & $\mathbf{44.0}$ & 74.4 & $1.69\times$ & 0.01 \\
    \qquad EchoPrime MViT & $\mathbf{6.34}$ & 7.34 & $1.16\times$ & 0.01 \\
    \qquad PanEcho & 0.273 & $\mathbf{0.231}$ & $0.85\times$ & 0.01 \\
    \addlinespace
    \multicolumn{5}{@{}l}{\textbf{\boldmath$256\times256$}} \\
    \multicolumn{5}{@{}l}{\quad Frame-level (FD $\downarrow$)} \\
    \qquad EchoCLIP & $\mathbf{0.128}$ & 0.175 & $1.37\times$ & 0.01 \\
    \qquad EchoFine & 4.78 & 4.68 & $0.98\times$ & 0.77 \\
    \qquad EchoPrime view & 0.518 & 0.485 & $0.94\times$ & 0.50 \\
    \qquad Inception & $\mathbf{33.9}$ & 41.3 & $1.22\times$ & 0.01 \\
    \multicolumn{5}{@{}l}{\quad Clip-level (KD $\downarrow$)} \\
    \qquad EchoJEPA & $\mathbf{17.0}$ & 35.9 & $2.11\times$ & 0.01 \\
    \qquad EchoPrime MViT & $\mathbf{2.19}$ & 4.91 & $2.24\times$ & 0.01 \\
    \qquad PanEcho & $\mathbf{0.197}$ & 0.300 & $1.53\times$ & 0.01 \\
    \addlinespace
    \multicolumn{5}{@{}l}{\textbf{\boldmath$320\times320$}} \\
    \multicolumn{5}{@{}l}{\quad Frame-level (FD $\downarrow$)} \\
    \qquad EchoCLIP & $\mathbf{0.108}$ & 0.130 & $1.21\times$ & 0.01 \\
    \qquad EchoFine & 6.74 & $\mathbf{4.64}$ & $0.69\times$ & 0.01 \\
    \qquad EchoPrime view & 0.644 & $\mathbf{0.479}$ & $0.74\times$ & 0.01 \\
    \qquad Inception & 34.2 & $\mathbf{30.4}$ & $0.89\times$ & 0.01 \\
    \multicolumn{5}{@{}l}{\quad Clip-level (KD $\downarrow$)} \\
    \qquad EchoJEPA & $\mathbf{16.6}$ & 20.3 & $1.23\times$ & 0.01 \\
    \qquad EchoPrime MViT & $\mathbf{1.90}$ & 2.32 & $1.22\times$ & 0.01 \\
    \qquad PanEcho & $\mathbf{0.169}$ & 0.331 & $1.95\times$ & 0.01 \\
    \bottomrule
    \end{tabular*}
    \par\vspace{2pt}
    \begin{minipage}{\linewidth}
    \footnotesize
    Each model generates one clip per CAMUS patient and view (500 patients, A2C and A4C) from the end-diastolic frame, with the ECG and frame timing of an EchoXFlow clip of the same view; the reference set is the non-end-diastolic frames and the ED-to-ES clips of the same patients. One model per domain and resolution (the fold-4 models), each scored over both views, hence no spread. Ratio: scan-converted / beamspace (above 1 favors beamspace). Bold: the better domain where $p_\mathrm{Holm}<0.05$.\\
    $p$: patient-level paired bootstrap, two-sided, Holm-adjusted across all 21 comparisons.
    \end{minipage}
\end{table}

\clearpage
\subsection{Foundation-model ejection fraction against 3D mesh volumes}\label{sec:foundation-model-ef}
The time-resolved 3D LV meshes give the ejection fraction (EF): the largest and smallest closed mesh volume of the cycle. We used it as a reference for the zero-shot EF estimates of EchoCLIP~\cite{christensen_visionlanguage_2024}, PanEcho~\cite{holste_complete_2025} and EchoPrime~\cite{vukadinovic_comprehensive_2026}, applied to the scan-converted A2C and A4C recordings of the 195 examinations that have both views and a valid mesh EF, without any fine-tuning. EchoCLIP and PanEcho estimate EF per view and we average the two; EchoPrime uses its own view-weighted study retrieval. \Cref{tab:foundation-model-ef} reports the agreement. All three models reach a two-view mean absolute error between 6.6 and 9.0 percentage points, in line with their published external validations, while PanEcho and EchoPrime overestimate EF by 7 to 9 points relative to the mesh volumes; after bias correction, their errors fall to 4.3 and 4.4 points.

\begin{table}[ht]
    \centering
    \caption{Zero-shot left-ventricular ejection fraction of public echocardiography foundation models (EchoCLIP~\cite{christensen_visionlanguage_2024}, PanEcho~\cite{holste_complete_2025}, EchoPrime~\cite{vukadinovic_comprehensive_2026}) against the 3D LV mesh volumes of 195 EchoXFlow examinations with both apical views.}
    \label{tab:foundation-model-ef}
    \footnotesize
    \setlength{\tabcolsep}{4pt}
    \begin{tabular*}{\linewidth}{@{\extracolsep{\fill}}ll*{5}{r}@{}}
    \toprule
    \textbf{Model} & \textbf{View} & \textbf{MAE [95\% CI]} $\downarrow$ & \textbf{RMSE} $\downarrow$ & \textbf{Bias} & \textbf{$r$} $\uparrow$ & \textbf{MAE$_\text{corr}$} $\downarrow$ \\
    \midrule
    EchoCLIP & A2C & 8.2 [7.4, 9.0] & 10.1 & +1.1 & 0.53 & 8.2 \\
     & A4C & 7.4 [6.6, 8.2] & 9.4 & $-$1.2 & 0.58 & 7.3 \\
     & Two views & \textbf{6.6} [5.9, 7.3] & \textbf{8.4} & \boldmath$-$\textbf{0.1} & 0.62 & 6.6 \\
    \addlinespace
    PanEcho & A2C & 7.9 [7.2, 8.7] & 9.5 & +7.2 & 0.69 & 4.9 \\
     & A4C & 8.0 [7.3, 8.8] & 9.5 & +7.1 & 0.74 & 4.9 \\
     & Two views & 7.7 [7.1, 8.4] & 9.0 & +7.1 & \textbf{0.77} & \textbf{4.3} \\
    \addlinespace
    EchoPrime & A2C & 9.6 [8.9, 10.3] & 10.8 & +9.3 & 0.74 & 4.3 \\
     & A4C & 8.4 [7.7, 9.2] & 9.9 & +7.4 & 0.69 & 5.1 \\
     & Two views & 9.0 [8.3, 9.6] & 10.2 & +8.5 & 0.74 & 4.4 \\
    \bottomrule
    \end{tabular*}
    \par\vspace{2pt}
    \begin{minipage}{\linewidth}
    \footnotesize
    EF in percentage points. MAE with a 95\% confidence interval; bias is prediction minus mesh; $r$ is Pearson's correlation; MAE$_\text{corr}$ is the leave-one-out bias-corrected MAE. 
    \end{minipage}
\end{table}
\clearpage

\newpage

\section{Visualization of 2D Speckle-Tracking Strain}
\begin{figure}[ht]
    \centering
    \includegraphics[width=0.99\linewidth]{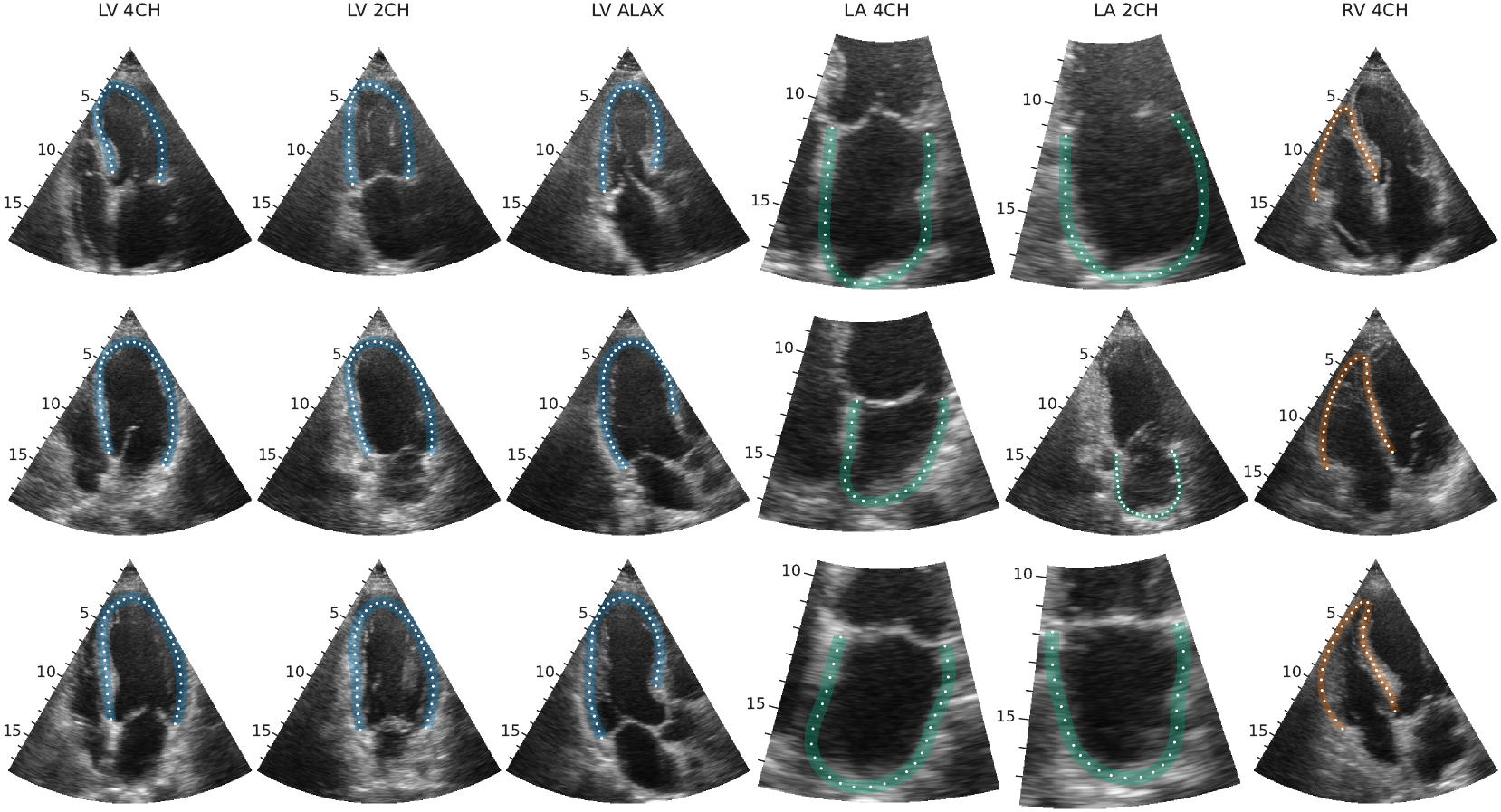}
    \caption{Three rows from three unique patients in EchoXFlow, each column corresponds to an echocardiographic view. LV tracings are more abundant in EchoXFlow compared to LA and RV annotations.}
    \label{fig:strain-images}
\end{figure}
\begin{figure}[ht]
    \centering
    \includegraphics[width=\linewidth]{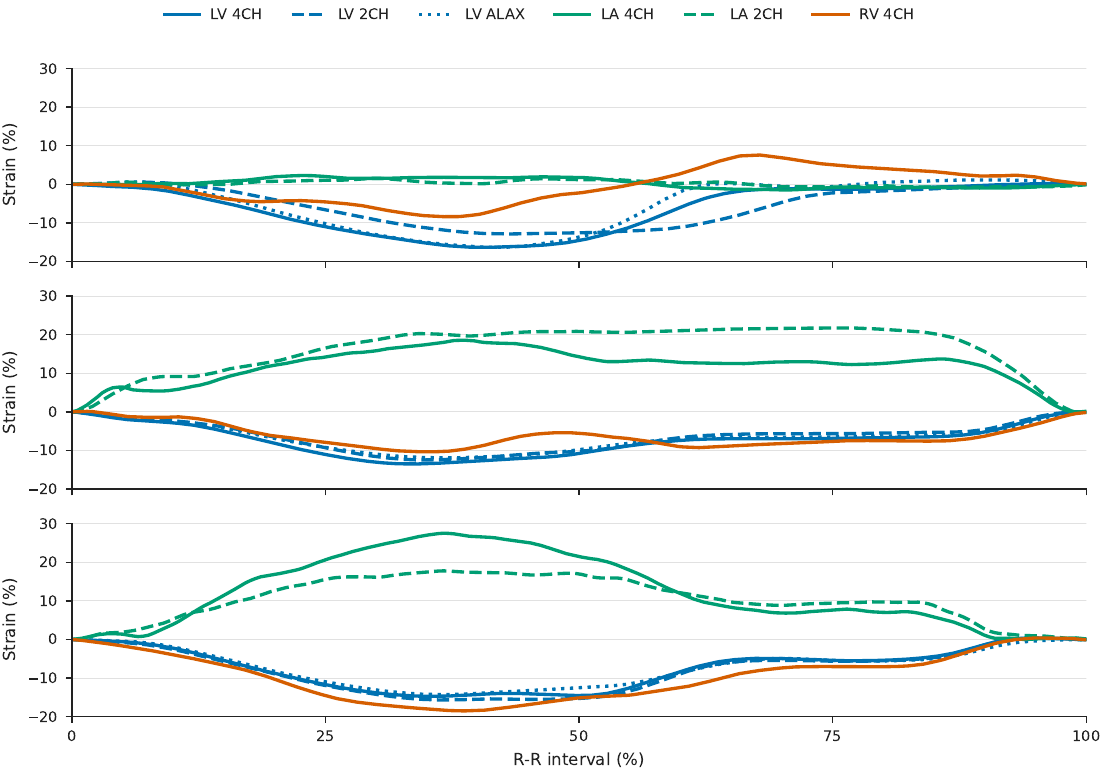}
    \caption{Strain curves over time; each row corresponds to a row in \Cref{fig:strain-images}.}
    \label{fig:strain-curves}
\end{figure}
\newpage

\section{Visualization of 3D Endocardial Segmentation}

\begin{figure}[ht]
    \centering
    \includegraphics[width=\linewidth]{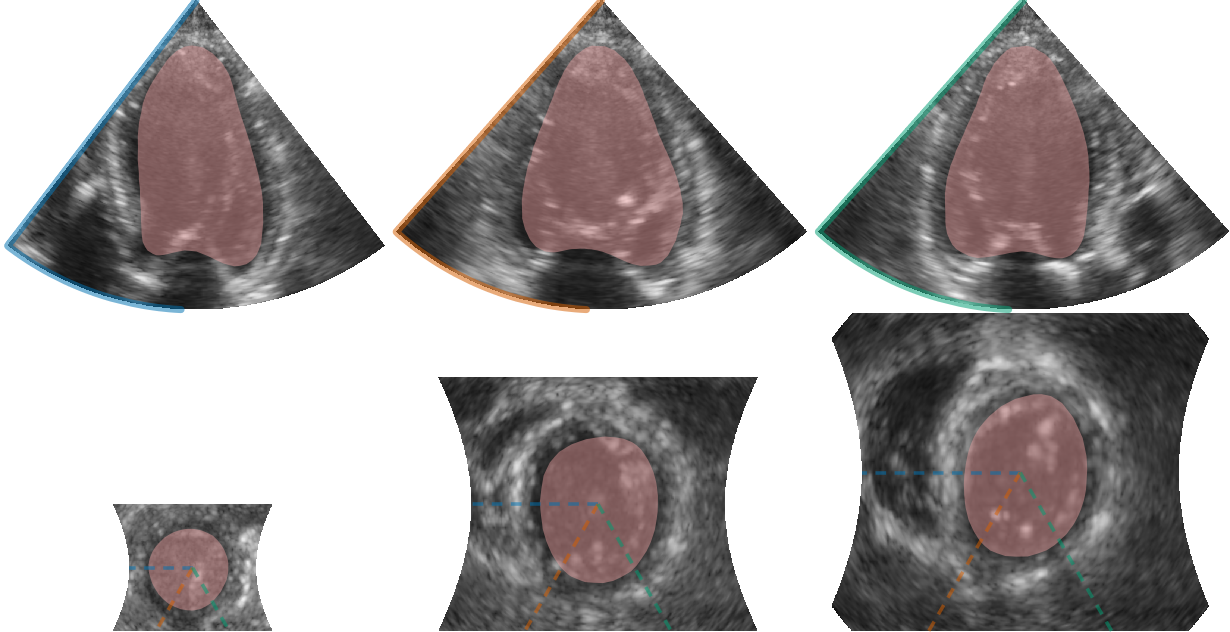}
    \caption{A single frame of a 3D volume with the LV endocardial segmentation mask. The top row shows slices of the 3D volume, separated by \SI{60}{\degree}, while the bottom row shows intersections approximately normal to the LV long axis at varying depth. Here, the top intersections roughly correspond to the A4C, A2C, and ALAX views, although the RV is not fully visible because the acquisition is LV-focused.”}
    \label{fig:volume-images}
\end{figure}
\begin{figure}[ht]
    \centering
    \includegraphics[width=\linewidth]{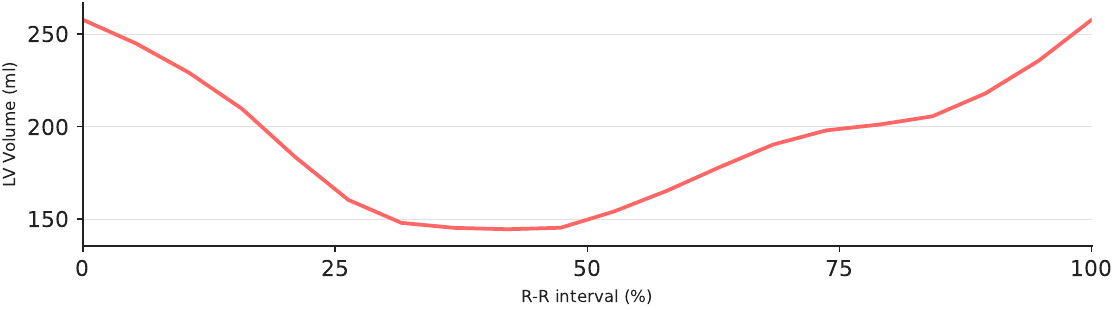}
    \caption{LV volume curve calculated from the endocardial segmentation mask in \Cref{fig:volume-images}.}
    \label{fig:volume-curve}
\end{figure}
\newpage

\section{View availability}
\begin{table}[ht]
\centering
\caption{Recordings per view. Other includes A5C, subcostal, and inferior vena cava. Recordings are labeled uncertain when the view classifiers disagree (EchoPrime and EchoFine).}
\label{tab:view-stats}
\footnotesize
\setlength{\tabcolsep}{3.5pt}
\begin{tabular*}{\linewidth}{@{\extracolsep{\fill}}lrrrrrrrrr@{}}
\toprule
\textbf{Statistic} & \textbf{All} & \textbf{A2C} & \textbf{A3C} & \textbf{A4C} & \textbf{PLAX} & \textbf{PSAX} & \textbf{Other} & \textbf{Bi/triplane} & \textbf{Uncertain} \\
\midrule
\multicolumn{10}{@{}l}{\textbf{All source recordings}} \\
\quad Total recordings, $n$ & \num{35377} & \num{1999} & \num{2047} & \num{9835} & \num{5139} & \num{5955} & \num{4553} & \num{1516} & \num{4333} \\
\quad Median per exam. & 52 & 3 & 3 & 15 & 8 & 9 & 7 & 3 & 5 \\
\quad IQR per exam. & 47--60 & 2--3 & 3--3 & 12--17 & 6--9 & 8--10 & 5--9 & 1--3 & 3--9 \\
\addlinespace
\multicolumn{10}{@{}l}{\textbf{Color Doppler subset}} \\
\quad Total recordings, $n$ & \num{8451} & \num{501} & \num{645} & \num{2119} & \num{1393} & \num{1956} & \num{993} & \num{45} & \num{799} \\
\quad Median per exam. & 12 & 1 & 1 & 3 & 2 & 3 & 1 & 0 & 1 \\
\quad IQR per exam. & 10--14 & 0--1 & 1--1 & 2--4 & 2--2 & 2--4 & 1--2 & 0--0 & 0--2 \\
\addlinespace
\multicolumn{10}{@{}l}{\textbf{Tissue Doppler subset}} \\
\quad Total recordings, $n$ & \num{1194} & \num{7} & \num{4} & \num{995} & \num{0} & \num{1} & \num{31} & \num{0} & \num{156} \\
\quad Median per exam. & 2 & 0 & 0 & 1 & 0 & 0 & 0 & 0 & 0 \\
\quad IQR per exam. & 0--3 & 0--0 & 0--0 & 0--3 & 0--0 & 0--0 & 0--0 & 0--0 & 0--0 \\
\bottomrule
\end{tabular*}
\par\vspace{2pt}
\begin{minipage}{\linewidth}
\footnotesize
Doppler subsets count recordings with at least two Doppler frames.
\end{minipage}
\end{table}

\section{Echocardiographic indices}
\begin{table}[ht]
\centering
\caption{Main annotation types in EchoXFlow. Detailed definitions, acquisition procedures, and measurement conventions are described in echocardiographic guidelines~\cite{mitchell_guidelines_2019, lang_recommendations_2015, galderisi_standardization_2017}.}
\label{tab:clinical_annotations}
\small
\setlength{\tabcolsep}{2.2pt}
\renewcommand{\arraystretch}{1.15}
\begin{tabular}{
>{\raggedright\arraybackslash}p{3.4cm}
>{\raggedright\arraybackslash}p{5.1cm}
>{\raggedright\arraybackslash}p{5cm}
}
\toprule
\textbf{Annotation type} &
\textbf{Source acquisition} &
\textbf{What it provides} \\
\midrule

2D strain contours
 & 2D B-mode; mainly A2C, A4C, ALAX
 & Time-resolved chamber contours, LV, LA, and RV, for deformation analysis\\
 \addlinespace

3D LV meshes
 & 3D B-mode; apical full-volume LV
 & Time-resolved inner surface mesh of LV, for volumetric assessment \\

 \addlinespace
Linear measurements
 & 2D B-mode or M-mode; mainly PLAX
 & Including LV cavity width, wall thickness, valve or vessel diameters \\

 \addlinespace
Tissue velocity 
 & TDI; mainly mitral and tricuspid annuli
 & Local heart-muscle motion graph with annotated landmarks, including e', a', s' \\

 \addlinespace
Blood-flow velocity 
 & PW/CW Doppler; A4C, A5C, ALAX
  & Blood-flow velocities over valves/outflow tracts/vessels\\

 \addlinespace
Color-flow regions
 & Color Doppler; valve/septal/vessel views
 & Spatial blood-flow maps used to localise regurgitation, turbulence, and shunts \\

 \addlinespace
 Doppler sample locations
 & PW/CW Doppler metadata
 & Doppler region of interest as anatomical locations or beam lines \\
 
 \addlinespace
Spectral Doppler envelopes
 & PW/CW Doppler traces
 & Outlined velocity curves used for peak velocity, timing, and flow measurements \\

\bottomrule
\end{tabular}

\par\vspace{2pt}
\begin{minipage}{\linewidth}
\footnotesize
\textit{Abbreviations.}
A2C: apical two-chamber view;
A4C: apical four-chamber view;
ALAX: apical long-axis view;
B-mode: brightness mode;
CW: continuous-wave;
LA: left atrium;
LV: left ventricle;
M-mode: motion mode;
PLAX: parasternal long-axis view;
PW: pulsed-wave;
RV: right ventricle;
TDI: tissue Doppler imaging.
\end{minipage}
\end{table}

\newpage

\clearpage
\section{Training and Validation Loss Curves}
\label{app:loss-curves}

\begin{figure}[!htbp]
    \centering
    \includegraphics[width=\linewidth]{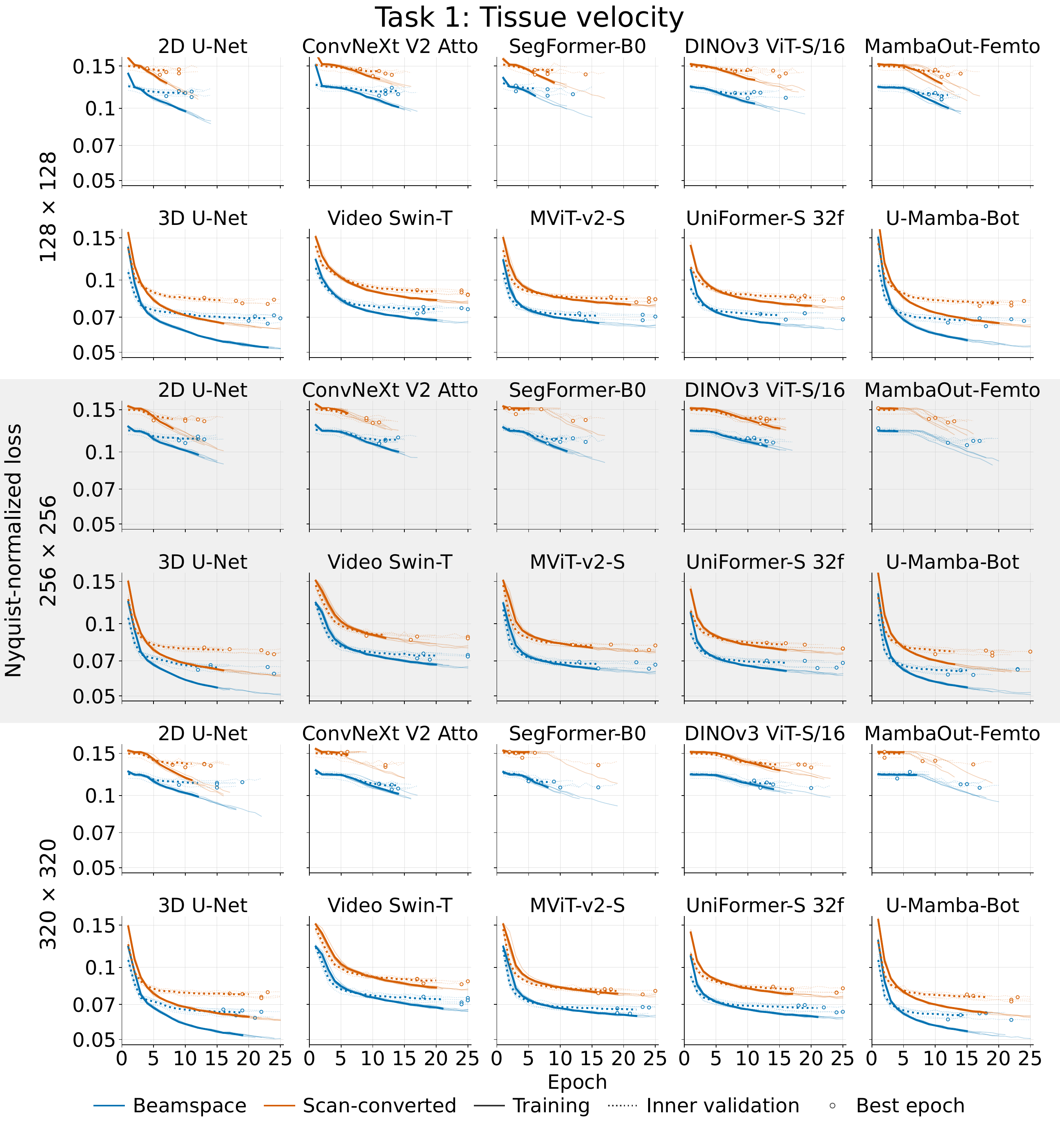}
    \caption{Training and inner-selection loss for tissue-velocity estimation (Task~1). For each resolution, the bottom row comprises time-aware models. The objective is the
    Nyquist-wrapped $L_1$ loss.}
    \label{fig:loss-curves-tissue}
\end{figure}

\begin{figure}[!htbp]
    \centering
    \includegraphics[width=\linewidth]{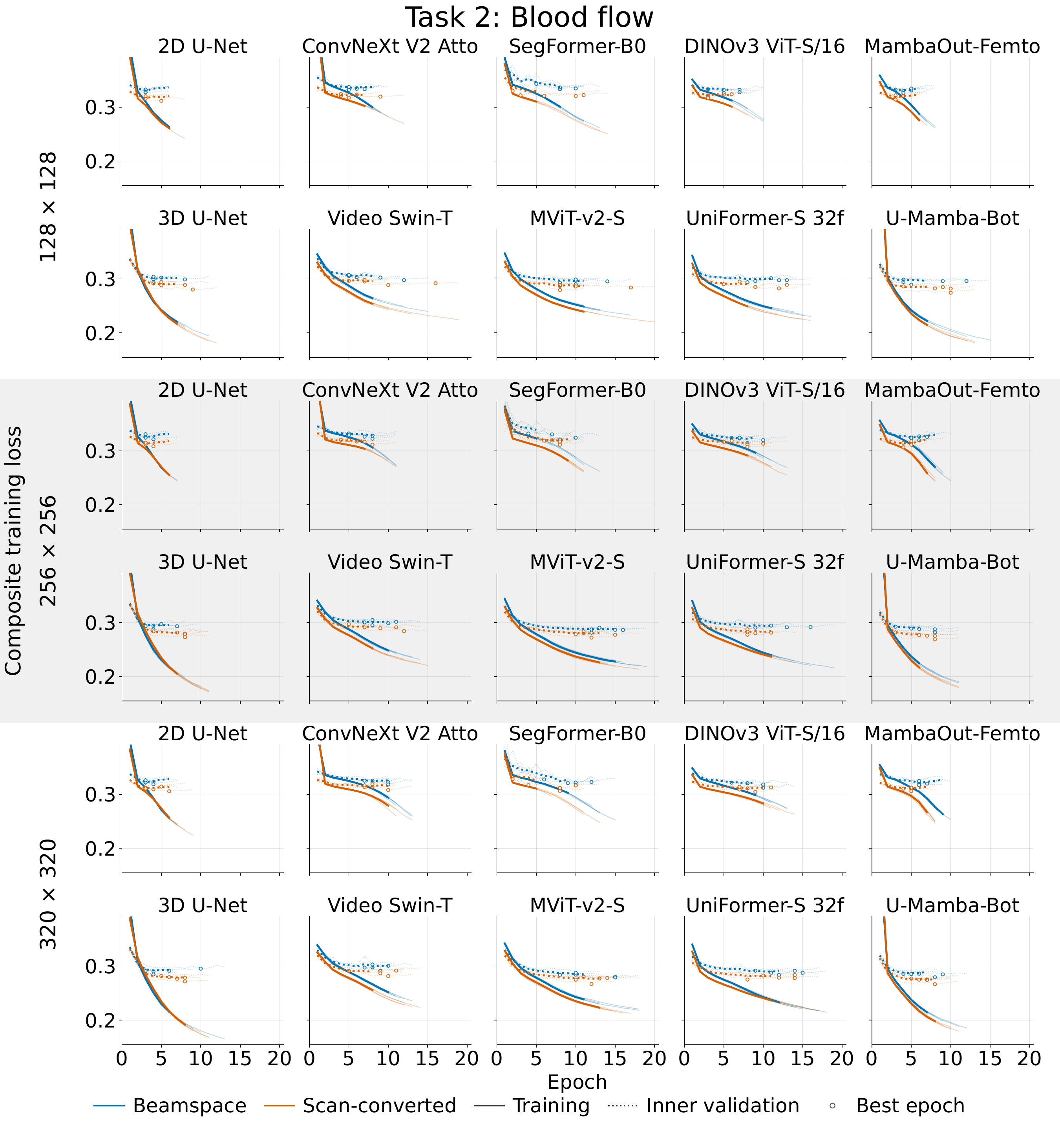}
    \caption{Training and inner-selection loss for blood-flow estimation (Task~2). The objective combines the
    velocity, power, and local velocity-variation terms.}
    \label{fig:loss-curves-color}
\end{figure}

\begin{figure}[!htbp]
    \centering
    \includegraphics[width=\linewidth]{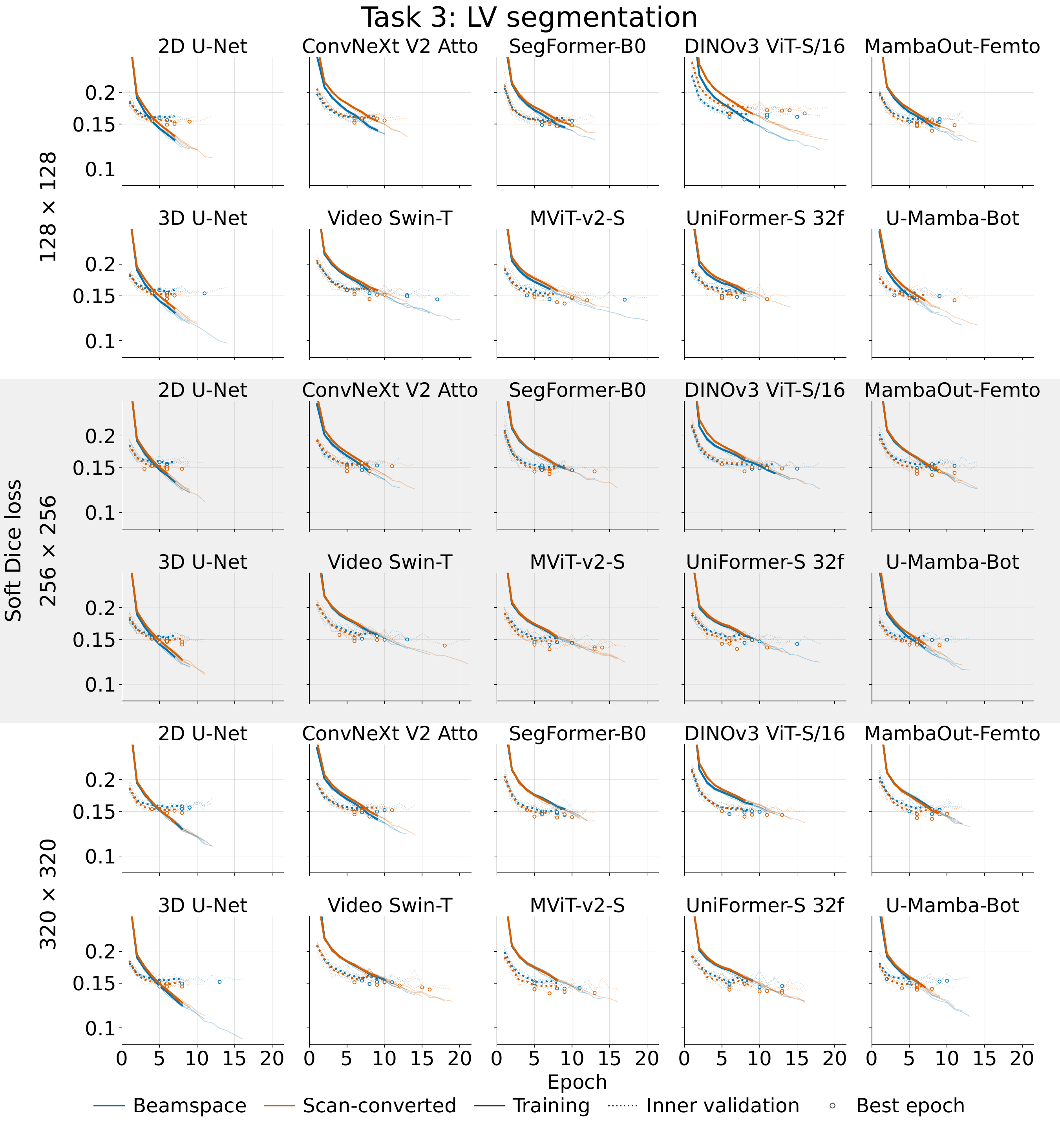}
    \caption{Training and inner-selection loss for left-ventricular segmentation (Task~3). The plotted objective
    is the foreground soft Dice loss, with the training-domain weighting described in \Cref{app:training-setup}.}
    \label{fig:loss-curves-segmentation}
\end{figure}

\begin{figure}[!htbp]
    \centering
    \includegraphics[width=\linewidth]{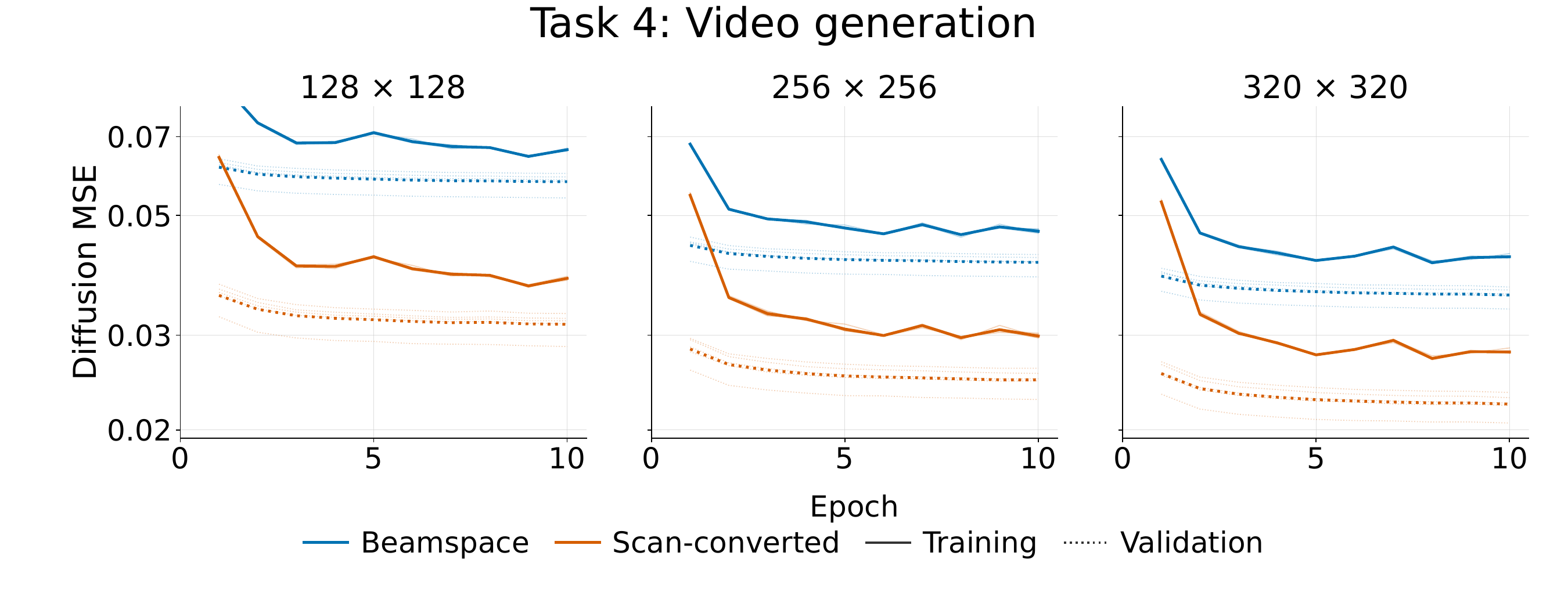}
    \caption{Training and validation loss for ECG-conditioned video generation (Task~4), over ten epochs and
    five folds. The objective is the diffusion $v$-prediction MSE. Because domains differ, lower MSE does not imply better fidelity when two models are compared in the scan-converted domain. Generated-video quality is assessed separately by feature distances in the scan-converted domain.}
    \label{fig:loss-curves-diffusion}
\end{figure}

\Cref{fig:loss-curves-tissue,fig:loss-curves-color,fig:loss-curves-segmentation,fig:loss-curves-diffusion}
show the recorded training and validation objectives at $128\times128$, $256\times256$, and $320\times320$ pixels. For Tasks~1--3, validation denotes the inner selection split used for early stopping; the reported outer evaluation fold is excluded from checkpoint selection. These curves display objectives in each training representation, which are not directly comparable across input representations. The cross-domain performance comparisons in the benchmark tables use the common Cartesian evaluation grid.

\clearpage

\clearpage

\clearpage

\end{document}